# Title

Evaluation of the impact of expert knowledge: How decision support scores impact the effectiveness of automatic knowledge-driven feature engineering (aKDFE)

***Authors:***

PhD student Olof Björneld [1,2,3,*] 0000-0002-8370-2950

Associate Professor Tora Hammar [1,2] 0000-0003-1549-2469

PhD Daniel Nilsson[4] 0000-0002-0354-8130

PhD Alisa Lincke [1,2] 0000-0001-9062-1609

Professor Welf Löwe [1,2] 0000-0002-7565-3714

***Affiliations***:

1. eHealth Institute, Department of Medicine and Optometry, Linnaeus University, S-391 82 Kalmar, Sweden.

2 Data Intensive Sciences and Applications (DISA-IDP), Department of Computer science and Media technology (CM), Faculty of Technology, Linnaeus University, S-391 82 Kalmar, Sweden.

3. Business Intelligence, IT Division, Region Kalmar County, S-392 32 Kalmar, Sweden.

4. AI Sweden, Sweden.

* Corresponding author:

E-mail addresses: olle.bjorneld@lnu.se (Olof Björneld), olle.bjorneld@regionkalmar.se (Olof Björneld)

## Abstract

Adverse Drug Events (ADEs), i.e., harmful, or unintended effects of a medication that occur during its proper use, such as side effects and allergic reactions, pose a significant challenge in healthcare as they impact patient safety and increase costs. Effective ADE detection from health data is important but often hindered by time-consuming manual processes in traditional Knowledge Discovery in Databases (KDD). This study addresses these limitations by investigating automatic Knowledge-Driven Feature Engineering (aKDFE), a method for patient-centric transformation of Electronic Health Record (EHR) data.

This study assesses aKDFE's effectiveness compared to automated event-based KDD that incorporates domain knowledge. It specifically investigates how incorporating domain-specific ADE risk scores for prolonged heart QT interval—extracted from the Janusmed Riskprofile (Janusmed) Clinical Decision Support System (CDSS)—affects ADE prediction performance using EHR data and medication handling events.

The study's findings indicate that, while aKDFE step 1 (event-based feature generation) alone did not significantly improve ADE prediction performance, aKDFE step 2 (patient-centric transformation) enhances the prediction performance. High Area Under the Receiver Operating Characteristic curve (AUROC) values in generated models suggest a strong correlation between some input features and the output variable, which aligns with the predictive power of patients' prior healthcare history for ADEs. A redesign of the Machine learning pipeline used, including broader grid searches and more complex models, could potentially even increase AUROC at a higher computational cost.

Statistical analysis did not confirm that incorporating CDSS Janusmed risk scores into the model's feature set enhanced predictive performance. In contrast to that, the patient-centric transformation applied by aKDFE proved to be a highly effective feature engineering approach. Limitations include the study's focus on a single project, potential bias from machine learning pipeline methods, and the reliance on AUROC as the primary evaluation metric.

In conclusion, aKDFE significantly enhances ADE prediction compared to using event-based EHR data, particularly when patient-centric transformation is applied. The inclusion of Janusmed ADE risk scores and medication route of administration had no statistically significant effect on predictive performance. Future work will explore (i) attention-based methods, such as those used in Large Language Models (LLMs), event feature sequences, and (ii) automatic methods for incorporating domain knowledge into the aKDFE framework.

## 1 Introduction

*Adverse Drug Events* (ADEs) represent a significant burden on healthcare systems, leading to patient harm and increased costs. Traditional *knowledge discovery in databases* (KDD) methods for ADE detection within healthcare often require substantial manual effort from domain experts and data scientists, resulting in prolonged completion times and high resource consumption. Enhancing the accessibility of medical information for end-users and researchers could accelerate the development of information-driven healthcare transformation, not just for ADEs.

The temporal relationships inherent in registered medical drug treatments necessitate a comprehensive investigation into prediction models based on series of events and data from *Electronic Health Records* (EHR) data when evaluating ADEs. High-performing ADE prediction models can improve the effectiveness of *Clinical Decision Support Systems* (CDSSs).

Prior research has demonstrated that transforming event-based data into a *patient-centric data* format enhances the predictive performance of generated models. Studies have shown the effectiveness of an automatic *Knowledge-Driven Feature Engineering* (aKDFE) method compared to manual KDD processes.

However, the data concentration achieved by transforming feature sets from event to a patient-centric data (PCD) format through pivoting or aggregation, also known as *entity-centric transformation*, offers a vast range of possibilities. Given that the evaluated feature sets contained temporal information, the selected transformations and generated models should be suitable for such characteristics. To streamline the KDD process by minimising manual work and human interaction, this study employs *N-grams* and *sum* values as operators for the transformation from event-based to patient-centric data format. The N-gram algorithm is widely used in various medical applications, but, to our knowledge, not in this specific context. We also briefly evaluate a *recursive neural network* (RNN) model.

Modern healthcare practices require access to the latest knowledge from *Clinical Decision Support Systems* (CDSSs). Many CDSSs are also called expert systems and utilise a knowledge base, and risk scores to provide recommendations. A key CDSS for ADEs in Sweden is the *Janusmed knowledge databases, Riskprofile is one of them*.

This study addresses the challenge of enhancing ADE detection from EHR data. To improve the effectiveness and efficiency of ADE detection, we investigate the impact of automatic Knowledge-Driven Feature Engineering (aKDFE), an entity-centric transformation method, on the performance of predictive models. These models utilise EHR data enriched with ADE risk scores from the Janusmed CDSS.

Specifically, we aim to answer the following research questions:

RQ1: Does aKDFE generate more effective ADE prediction models compared to automated event-based KDD with incorporated domain knowledge?

RQ2: Does incorporating domain-specific ADE risk scores from a CDSS improve the predictive performance of ADE prediction models?

We hypothesise that (H1) aKDFE will generate prediction models with higher classification performance in detecting ADEs from EHR data, compared to automated event-based KDD with incorporated domain knowledge, and (H2 prediction models incorporating ADE risk scores from a CDSS will demonstrate improved classification performance in detecting ADEs compared to models trained without such domain expert knowledge.

Our evaluation will assess these hypotheses, contributing to the development of more robust ADE detection methodologies.

# 2 Background

To provide a comprehensive understanding of how data science and machine learning techniques, such as aKDFE, are employed in ADE detection, the following background information is presented.

## 2.1 Electronic health records (EHR)

EHRs are fundamental tools in clinical healthcare. The reuse of the detailed data contained within EHRs is of significant interest, as highlighted in the proposed European Health Data Space regulation [1]. A primary challenge in utilising EHR data lies in its inherent unstructured, text-based format. This characteristic necessitates a collaborative effort between health care domain experts and data scientists to effectively transform the raw data into actionable knowledge [2].

As described by [3], *low-level* EHR data are typically stored in a relational database structure, characterised by *one-to-many* (*1:N*) cardinalities between entities such as patients, episodes, and visits. This relational structure, illustrated in Figure 1, demonstrates how a single patient can be associated with multiple episodes ($1{:}N_E$) and visits ($1{:}N_V$). Each episode can have multiple diagnoses ($N_E{:}N_{ED}$), and each visit can have multiple diagnoses ($N_V{:}N_{VD}$) etc. This illustrates the hierarchical *many-to-many* nature of EHR data, where a patient entity serves as the root, linked to subordinate entities such as episodes and visits, which in turn are linked to diagnoses.

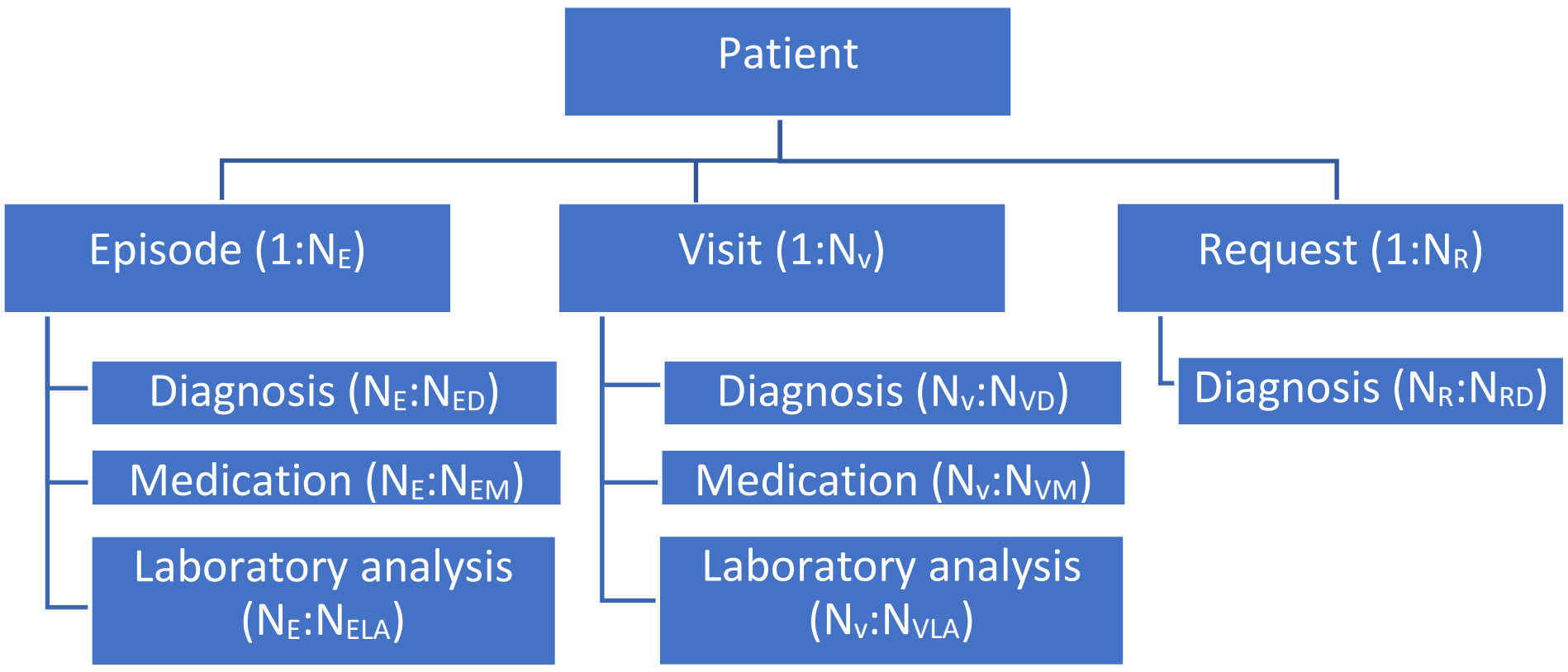

*Figure 1 Relationship of medical information in EHRs. 1:$N_x$ represent one-to-a-positive number of the entity x.*

## 2.2 Medications and drug consumption

Medications play a crucial role in daily healthcare treatment, significantly improving the lives of numerous patients. However, adverse drug events (ADEs) occur frequently and are a common source of negative outcomes, including patient suffering, unnecessary healthcare utilisation, and mortality. ADEs are defined as harm resulting from a medication itself or its inappropriate use [4, 5].

Patients receive medications based on prescribing decisions authorised by health care professionals. *Medication handling events* encompass prescribed medications dispensed at pharmacies, as well as those administered and documented within healthcare settings, such as during hospitalisation. Medication events often involve complex, parallel subprocesses and extend over prolonged periods. Consequently, the data required to explain and predict ADEs are typically distributed across extended timelines with low information density. To enhance the classification performance of ADE prediction models, generating informative features from "patient-specific factors", such as features in a patient-centric data format, is essential [6].

## 2.3 Clinical decision support system (CDSS)

CDSSs offer a promising avenue for preventing ADEs by identifying potentially harmful medication combinations, whether implemented in healthcare settings or pharmacies [4, 7]. While many current CDSSs rely on rule-based systems, necessitating manual monitoring and score updates, data-driven approaches, often leveraging Machine Learning, present a compelling alternative. In Sweden, the Janusmed knowledge database is a prominent resource for medication-related CDSSs; however, its adoption across Swedish regions remains limited, including its integration with EHRs [8, 9]. In systematic reviews regarding pregnancy care and nursing it is stated that daily health care has use of CDSSs. However, despite of many CDSSs in clinical use, there is a need for higher predictive effectiveness and better validation of CDSS models [10, 11].

## 2.4 Knowledge discovery

*Knowledge discovery (in databases)* (KDD) is the process of how to find novel and potentially useful knowledge in large databases [12]. The process typically follows a sequential methodology, encompassing (i) data load and transformation, (ii) extraction of useful information (also referred to as *data mining*), and (iii) evaluation of the discovered knowledge [13].

Numerous KDD methods exist, each tailored to specific contexts [14-16]. This study employs the KDD process introduced at the first KDD workshop [17], as illustrated in Figure 2. Throughout this paper, the terms "knowledge discovery" and KDD are used as interchangeable.

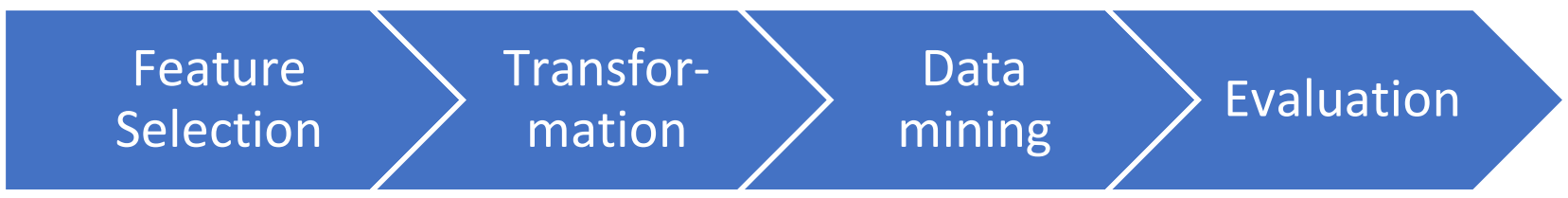

*Figure 2 The KDD process.*

*KDD* within the medical domain, utilising EHRs, requires the collaborative expertise of health care domain experts, data scientists, and statisticians. Through an iterative ideation process, these professionals refine low-level variables, eliminate irrelevant ones, and generate novel variables. In computer science, these *variables* are commonly termed *features*. Therefore, the term features will be employed consistently to describe variables throughout this study. This iterative, so-called, *ideation*, when successful, results in an enhanced feature set that can generate prediction models with high performance. The same process, with aimed at of condensing low-level data into informative features, is also known as *feature engineering* (FE) [18].

The iterative collaborative KDD process described above, previously referred to as *manual* KDFE (mKDFE), was introduced as a systematic but manual approach [3]. However, this manual nature requires significant input from domain experts and data scientists, who are often scarce resources. Consequently, the process is inherently tedious and time-consuming, extending over prolonged periods [3]. To mitigate these limitations, a *framework for the automation* of KDFE has been developed and evaluated [19].

## 2.5 Source data formats

An *entity-centric data* formatted feature set organises data such that each entity occupies a single row, with the columns representing the entity's features. The entity-centric transformation, which converts event-based data to an entity-centric data format, has been studied. For instance, the validity of using entity-centric transformation compared to time series analysis has been evaluated through Monte Carlo simulations. These simulations demonstrate that under specific conditions, comparable results can be achieved [20]. Furthermore, this study provides guidelines for determining whether a longitudinal survey approach is appropriate.

In this study, we adopt an *entity-centric data* format, as it offers greater accuracy compared to alternative formats, for representing electronic health record (EHR) feature sets. Consequently, we will refer to *patient-centric data format* and *patient-centric transformation* (PCT) rather than entity-centric data format and entity-centric transformation.

The patient-centric data format is characterised by a *one-row-per-patient* structure, where: (i) each row corresponds to a unique patient, (ii) a patient only exists in one row, and (iii) columns denote specific characteristics or features of that patient.

Health-related issues often manifest as progressive developments over time, resulting in event-based time series and a one-patient-to-many *(1:N)* structure.

These *event-based* health series present significant evaluation challenges due to their substantial data and computational resource requirements. Furthermore, the transformation of features from event-based formats into patient-centric data format necessitates careful consideration of feature transformation and aggregation techniques [21].

Typically, features extracted from EHRs during the initial KDD process are not in patient-centric data format. Instead, they are event-based, meaning each feature can be recorded across multiple rows (corresponding to distinct events) for a single patient. Additionally, EHR events exhibit a high degree of sparsity compared to traditional time series data. Table 1 illustrates these differences.

*Table 1 Dataset with temporal information in the health domain.*

| **Dataset type** | **Description** | **Example** |
|---|---|---|
| Time series | Continuous recordings of observations, typically sampled at a relatively high frequency. | Home blood pressure monitoring, or motion detectors for fragile patients. |
| Event series | Observations with associated multiple attributes, exhibit a distinct temporal pattern. Unlike time series registrations, event series are marked by their sporadic and random occurrence. | Non-deterministic medication handling events or emergency care visits. |

## 2.6 N-grams

*N-grams*, a widely established method, were first introduced over 70 years ago [22]. This technique involves concatenating *N* consecutive string groups, delimited by specified separators. In the medical domain, N-gram find applications in diverse areas, including protein sequencing [23], image classification [24], and text mining of EHRs [25]. While historically successful in the *Natural Language Processing* (NLP), N-grams have been increasingly superseded by more recent methodologies, notably *Large Language Models* (LLMs) [26]. However, LLMs present certain limitations, such as (i) a substantial requirement for text corpora and (ii) the need for advanced computing hardware. The former is particularly salient in the medical field, where extensive text datasets may be limited. Consequently, N-grams and K-means remain viable and effective alternatives, even when LLMs are not a feasible option [27].

## 2.7 Feature Engineering (FE), Feature Selection (FS), and Feature Stores

FE encompasses the construction, generation, and extraction of features, often leading to enhanced *prediction* outcomes (classification or regression). While resource-intensive, FE is most effective when health care domain and data experts collaborate closely [28, 29].

*Knowledge-driven feature engineering* (KDFE) refers to the iterative process of FE conducted in conjunction with domain experts [3]. This iterative approach aims to progressively improve the predictive power of the engineered feature set. In this study, features exhibiting high predictive power and informative features are used interchangeably. When KDFE involves manual transformation of new features, it is termed *manual* Knowledge-Driven Feature Engineering (mKDFE). Conversely, *automated* knowledge-driven feature engineering (aKDFE) describes the process when this transformation is automated.

FE aims to extract meaningful information, e.g., informative features from EHR. Common FE operators include, (i) deriving intrinsic data such as weekday, year, and geographical areas, (ii) aggregating features using statistical measures, (iii) applying unary operators such as sin(x), $x^2$, and $\sqrt{x}$ [30], and (iv) performing binary operations on features values such as sum, difference, product, and quotient.

Medical data frequently incorporates temporal information, leading to intricate relationships and a significant risk of causality-related bias. However, these temporal sequences hold substantial potential for generating informative features, especially when leveraging longitudinal patterns rather than isolated events or data points [31]. Indeed, as demonstrated in [31], FE based on temporal patterns yields highly informative features.

In parallel to aKDFE, alternative FE frameworks have emerged within the medical domain, specifically tailored for time-to-event outcomes [32]. Concurrently, general-purpose FE tools have been developed to facilitate broader applicability and user accessibility, often implemented within python environments. Notable examples include autofeat [33], iFeature [34], FIDDLE [35], and iLearn [36].

### 2.7.1 Automatic FE

*Automated FE systems* have been developed across diverse domains, including protein interaction prediction [30, 37] and energy consumption forecasting [38]. These systems generate novel features through either rule-based methodologies or fully automated processes, leveraging predefined operators. The automated feature generation can be optimised using genetic algorithms [30]. The specific operations employed vary depending on the data type of the feature, such as integer, datetime, or character [39, 40].

Iterative FE has been implemented through various methods. One notable approach leverages Reinforcement Learning (RL), where an agent is trained to maximise a goal function based on feature informativeness [41].

In the wake of the development of LLMs, sematic-oriented automatic FE techniques have emerged. Context-Aware Automatic FE (CAAFE) utilises LLMs to automatically extract informative features, leading to prediction models with enhanced performance [42].

Automated FE systems and models fall within the Automated Machine Learning (AutoML) domain. While AutoML holds significant potential in healthcare, further research is needed to improve efficiency when processing large retrospective datasets [43]. Given the unstructured and often text-based nature of Electronic Health Record (EHR) data, methods like CAAFE could prove particularly beneficial.

## 2.8 Machine learning models

Machine learning models are designed to suit specific prediction tasks and input feature types. For patient-centric data formatted feature sets, which lack temporal information, *feed forward* models are commonly employed. In this context, they are used to develop a prediction model for post-surgical cardiac risk [44].

The term "feedforward" describes the unidirectional flow of data through the model during training, from inputs through intermediate nodes to outputs. Recursive models, on the other hand, are typically utilised for prediction tasks involving features with inherent temporal dependencies. These models find applications in various domains, including speech recognition, event-series analysis, and anomaly detection [45].

Examples of recursive models include RNNs, *Long Short-Term Memory* (LSTM), and *Gated Recurrent Unit* (GRU) [45]. Although recursive models are designed for temporal data, studies suggest that FF models can achieve comparable performance. According to [46], a contributing factor is that recursive models, due to their intricate structure, often require substantial training datasets to yield high-performing prediction models, as illustrated in the prediction of dementia from EHR data.

## 2.9 Evaluation of machine learning-generated classification models

To assess the inherent knowledge contained within a feature set, employing a *machine learning pipeline* is an effective approach [47, 48]. The knowledge extracted from aKDFE feature sets can be evaluated through the generation of classification models. This model generation is facilitated by a machine learning pipeline, which encompasses a series of processes designed to produce a comprehensive and valid model. For detailed descriptions of the machine learning pipeline subfunctions, cf. the Appendix.

The primary objective of the machine learning pipeline is to generate trustworthy models. Trustworthiness is a critical concern within the broader AI domain, as underscored by the guidelines of the EU AI Act [49]. Table 2 presents key machine learning pipeline objectives, including specific aims related to trustworthy AI. These objectives are derived from our own experience and supported by the literature [50, 51].

*Table 2 Objectives for machine learning pipeline for prediction model generation.*

| Machine learning pipeline objectives | Detailed description |
|---|---|
| Effectiveness | Generate high-quality models with consistent and reliable performance. |
| High generalisability | Generated models should perform well on new data. |
| Efficiency | The pipeline should minimise manual work. |
| Scalability | The pipeline should handle increasing data volumes and processing demands. |
| Trustworthiness | Generated models that comply with fairness, transparency, and responsible development. |

## 2.10 Knowledge Representation and Measurement

A quantitative measure of knowledge that can be gained from data or feature sets is central to many domains and tasks, not just the medical domain. The compare the inherent knowledge within different feature sets, researchers compared relevant groups of features [52]. These include but are not limited to: (i) Shannon entropy, (ii) goodness of fit, and (iii) wavelet entropy [53, 54]. In the context of machine learning-based classification, (iv) the area under the receiver operating characteristic curve (*AUROC* or *ROC AUC*) is a particular significant metric [55].

For statistical hypothesis testing, numerous valid methods are available. One commonly used test is *analysis of variance* (ANOVA), which, for instance, has been applied to EHR data in the assessment of frailty index [56].

## 2.11 Automatic KDFE (aKDFE)

To mitigate the limitations of mKDFE, an automated version, aKDFE, has been developed and evaluated. aKDFE retains the foundational framework of mKDFE, as previously described [19].

The aKDFE framework, which has been presented and validated using clinical studies with multifactorial explanatory models [3, 19] consists of two primary steps: (i) step one - the generation of novel event-based features, and (ii) step two - the transformation of these features into a patient-centric data format.

In essence, KDFE leverages EHR data to construct informative features, leading to improved classification model performance compared to models utilising only EHR data. Aligning with the knowledge pyramid concept [57], by using KDFE, more knowledge can be discovered from less data in a shorter time and with fewer resources.

# 3 Methodology

This study compares the classification performance of ADE detection from various prediction models. These models leverage medication handling events, EHR data, and Janusmed risk scores encompassing event-based EHR data and feature sets generated by aKDFE in a patient-centric data format.

Experiments, involving the development of prediction models, were conducted using medication events and clinical outcomes extracted from EHR data. The performance of these models was assessed using quantitative metrics. The primary objective is to evaluate the predictive capability of feature sets generated by different FE methods, both with and without the inclusion of CDSS risk scores (Janusmed), through the creation and evaluation of classification models.

Key aspects of this study include: (i) the development of a strategy to evaluate diverse feature set input formats for machine learning models, (ii) the transformation of event-based features into a patient-centric data format, and (iii) the assessment of the impact of incorporating CDSS risk scores (Janusmed) within aKDFE.

A machine learning pipeline, as detailed in Section 2.10, was employed to evaluate the derived knowledge. This evaluation focused on the predictive power of selected features concerning clinical outcomes (ADEs). To specifically assess prediction models utilising feature sets with temporal information, a RNN model was developed.

## 3.1 Data sources

### 3.1.1 Janusmed Riskprofile database

The Swedish knowledge database Janusmed facilitates the estimation of risk scores for ADEs based on concurrent drug use. While Janusmed Riskprofile encompasses nine distinct ADE categories, this study focuses solely on the risk of prolonged QT interval [58].

### 3.1.2 Study related medication handling events

For each day of the ten-year study period, concurrent medications for each patient were determined based on prescription drugs dispensed at pharmacies and medications administered in healthcare settings, as recorded in the EHR within the preceding 120 days. These two event types are in this study referred to as medication handling events, cf. Section 2.2. The 120-day period was chosen because medications for continuous treatment are often collected approximately every three months, whereas medications administered in healthcare settings are recorded at the time of administration.

Medications were assigned a risk score, which was then aggregated using a predefined algorithm to determine an overall risk level for each patient. The risk levels were categorised as follows: level 0 (no increased risk), level I (somewhat increased risk), level II (moderately increased risk), and level III (significantly increased risk, the highest level). As part of this classification process, topical medications (based on medication route of administration) were excluded, and only unique medications were considered in the risk assessment. If a patient received the same substance multiple times within the 120-day exposure period, it was counted only once.

### 3.1.3 Clinical outcomes from EHR

The study outcomes consist of events or other registrations in the EHR that may indicate an ADE related to QT prolongation. The outcomes were identified by domain expert [58]. While the primary outcome of interest is a specific type of abnormal heart rhythm (Torsades de Pointes, TdP), additional outcomes were included to capture potential cases.

For each outcome category, it was assessed whether an event occurred within 365 days following the index date (yes/no), the number of days until the first occurrence, and the total number of recorded outcomes. Additionally, for the analysis, data on prior diagnoses and days of hospitalisation during the year preceding the index date were collected.

The data were stored in an event-based format, meaning that each patient can exist on many rows for each type of event, e.g., clinical outcome.

### 3.1.4 Predicted outcome

In this study, the binary output variable, denoted as 'Y', represents the occurrence of a clinical outcome (ventricular arrythmia) indicative of QT interval prolongation (ADE). 'Y' takes a value of 1 if the patient experienced the ADE, and 0 otherwise.

## 3.2 Feature set generation

The collected medical events, according to the medical registry study's requirements, serve as the foundation for a typical medical research project. In this study, this foundation is represented by the 1-EVENT feature set. A more refined project, incorporating domain knowledge through CDSS scores, is represented by the 2-EVENT feature set. The 1-EVENT-aKDFE and 2-EVENT-aKDFE feature sets are derived from the step one of the aKDFE process, while the 1-PCD-aKDFE and 2-PCD-aKDFE feature sets are derived from both the step one and two. For detailed information regarding the feature sets and their associated groups, see Table 3.

*Table 3 Included feature sets in the study.*

| Feature set name | Data format type | Feature set group | Feature track, w./wo. use of Janusmed |
|---|---|---|---|
| 1-EVENT | Event | Event EHR | Without |
| 2-EVENT | Event | Event EHR | With |
| 1-EVENT-aKDFE | Event | Event aKDFE | Without |
| 2-EVENT-aKDFE | Event | Event aKDFE | With |
| 1-PCD-aKDFE | PCD | PCD aKDFE | Without |
| 2-PCD-aKDFE | PCD | PCD aKDFE | With |

This research evaluates the performance of three distinct feature set groups across two experimental tracks: with and without the application of Janusmed. Figure 3 summarises the characteristics of these feature sets, including their size and number of features. Figure 4 provides a visual representation of the feature generation workflow for each set.

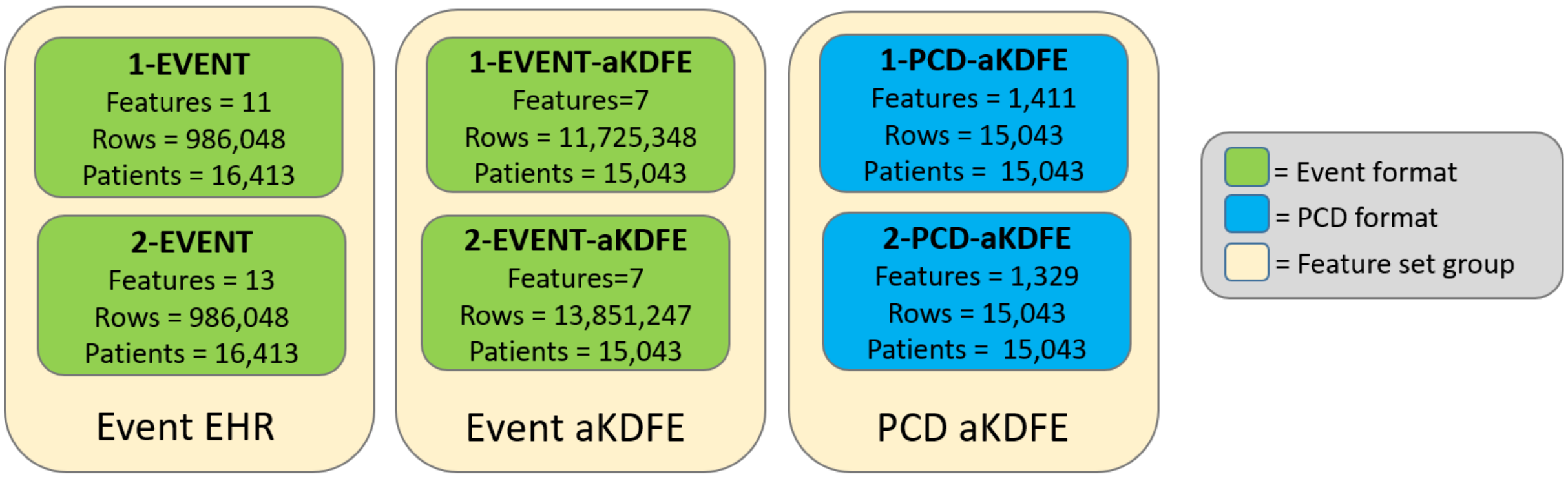

*Figure 3 Included feature set with associated groups.*

The overall workflow for the generation of feature sets is depicted in Figure 4.

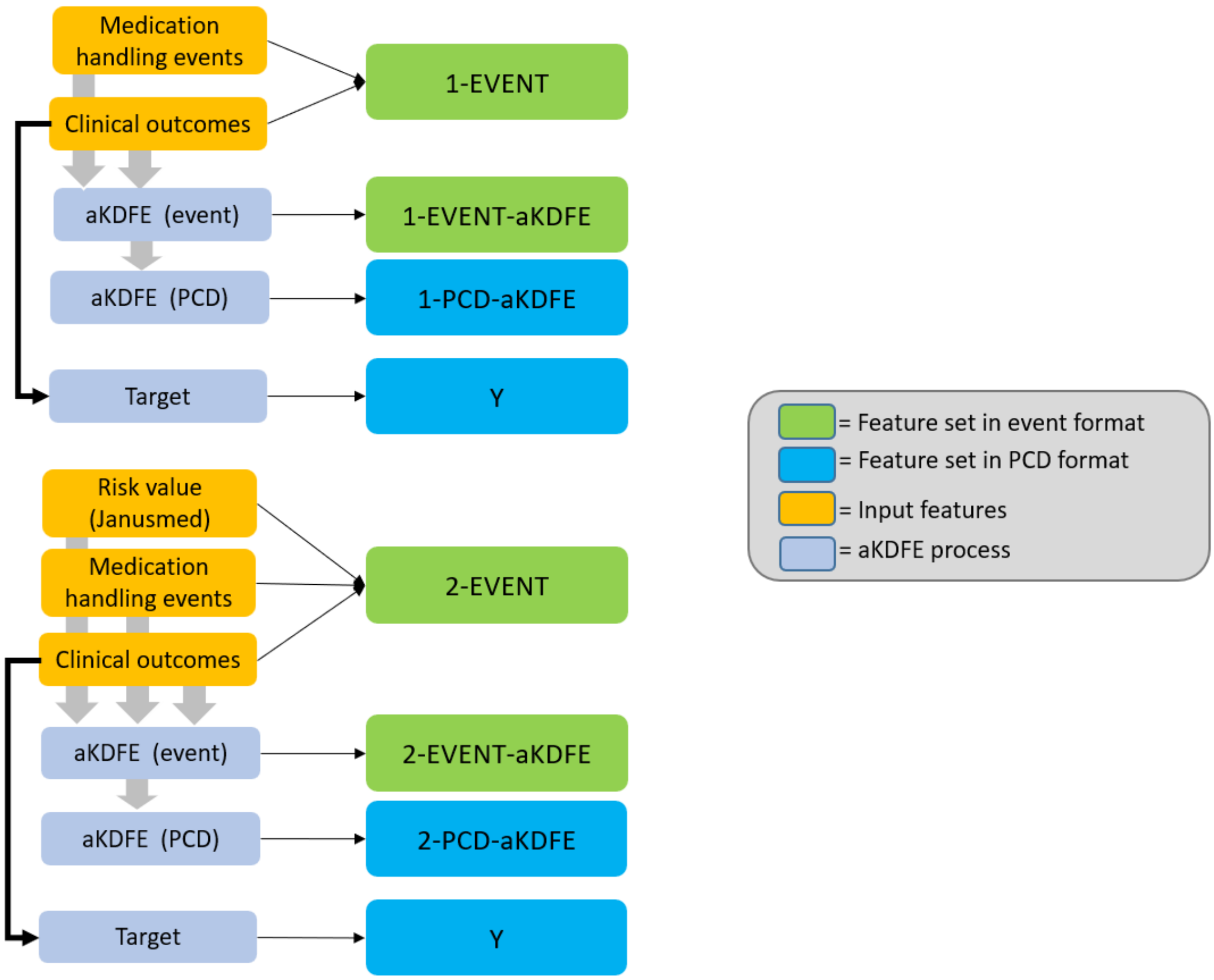

*Figure 4 Schematic workflow for the generation of the included feature sets.*

### 3.2.1 Collected data structure

From own experience, medical research projects exhibit substantial heterogeneity, resulting in diverse structures and information content within collected and generated health data. Consequently, data alignment and transformation are often necessary to conform to a standardisedaKDFE process. To accommodate the varied nature of health data, features are represented using decimal, character, or date data types, as detailed in the Appendix. These datasets are designed to capture data in an event-based format, mirroring the structure of real-world EHR data. However, for the development of machine learning-based prediction models, the patient-centric data format is preferred [21].

### 3.2.2 Entity-centric transformation

To align with specific medical research questions, event-based data is often transformed into a patient-centric data format. This transformation involves aggregating or pivoting data such that each patient is represented by a single, unique row. In this study, we refer to this process as *entity-centric transformation*. Prior research indicates that features derived from this transformation yield models with enhanced predictive performance [3].

## 3.3 Feature engineering by aKDFE

The aKDFE process comprises two sequential steps: (i) step one - generation of event-based aKDFE feature sets from EHR data, and (ii) step-two patient-centric transformation of these event-based feature sets into a patient-centric data format. This study employs two parallel experimental tracks to evaluate aKDFE performance: (i) without the incorporation of Janusmed risk scores, and (ii) with the incorporation of Janusmed risk scores.

### 3.3.1 Step one aKDFE - generation of event-based features

Figure 5 illustrates the step one aKDFE feature generation. The aKDFE framework facilitates the transformation of event-based EHR feature sets into event-based aKDFE feature sets. The specific transformations implemented within the aKDFE framework were derived from prior health research projects and constitute a core component of the methodology. Previous analysis and evaluation, as detailed in [3], demonstrated the superior effectiveness of aKDFE compared to a baseline approach.

Each subprocess within aKDFE generated a substantial number of features, many exhibiting low coverage and importance. Consequently, a rank-based feature selection was performed after each subprocess, as depicted in Figure 5. This ranking was based on feature coverage across unique patients.

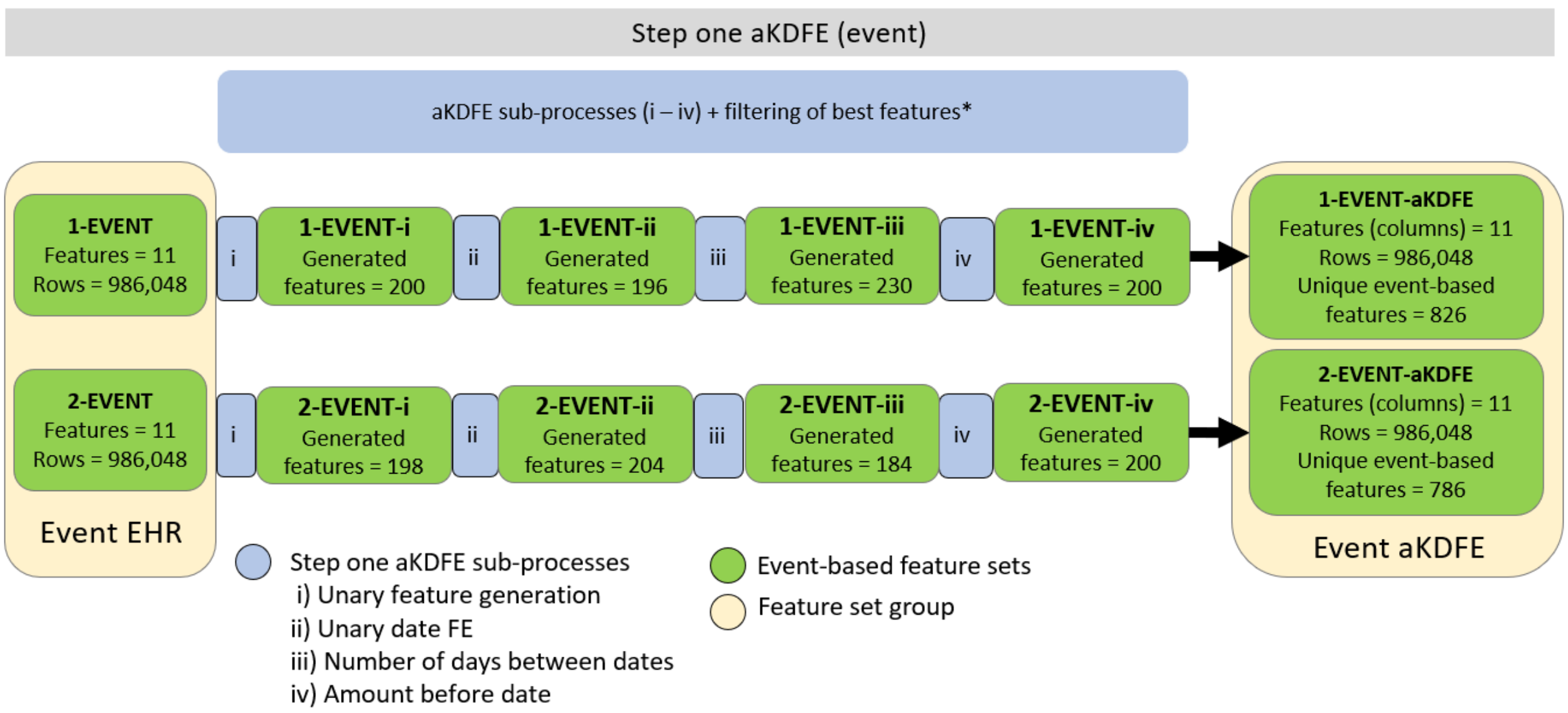

*Figure 5 Generation of aKDFE-event features sets in step one of aKDFE.*
** The 200 features with the highest patient coverage were selected following each subprocess and used as input for the subsequent subprocess. In cases where multiple features shared the coverage ratio of the 200th-ranked feature, all such features were included. Some subprocesses did not generate 200 features.*

Figure 5 details the aKDFE process, outlining the methods employed by each subprocess. Feature transformations are implemented using the data warehouse (DWH) tool SAP Data Integrator [59]. The resulting outputs, including feature metadata, generated features, and feature values, are stored within a relational database and the aKDFE framework, both residing in a Microsoft SQL database.

### 3.3.2 Step two aKDFE - Patient-centric transformation of events in the aKDFE feature sets

The transformation of event-based aKDFE feature sets into patient-centric data formatted aKDFE feature sets was achieved through two distinct patient-centric transformation methods: (A) N-gram generation on ordered events, and (B) summation of feature values.

A) N-gram generation

Events were ordered by patient identifier, feature event timestamp, and feature id. Subsequently, N-gram features were generated following the methodology described in Section 2.6. The maximum number of N-grams per patient is constrained by the total number of events associated with that patient (*N-gram-max*). Consequently, theoretically, *N* can range from *1* to *N-gram-max* for each patient. This study prioritises methodological evaluation over optimal predictive performance; therefore, only *N=1* was implemented and evaluated. However, employing small *N* values would have introduced non-deterministic issues due to the prevalence of multiple events with identical timestamps (*Patient-same-Event*). To mitigate this, *N* must exceed the maximum *Patient-same-Event* value observed across all patients. Following N-gram generation, the total count of N-grams for each patient was calculated and stored as a feature within the patient-centric data formatted aKDFE feature sets.

B) Sum of feature values

The second method involved summing the stored feature values for each unique event within the event-based aKDFE feature sets. These feature values originated from step one aKDFE (feature generation), cf. 3.3.1.

## 3.4 Metadata extraction

Prior to the initiation of an aKDFE-enhanced project, knowledge acquired from previous projects can be integrated into the aKDFE framework [19]. Upon project completion, the accumulated aKDFE metadata can be generalised into structured knowledge and incorporated into the framework. These updates facilitate continuous learning, enhancing the performance of subsequent aKDFE-driven projects.

Each aKDFE iteration yields new features, which are stored alongside their calculated values. The aKDFE framework automatically extracts valuable intrinsic feature metadata, including data type, complexity, construction, and provenance. Component elements of generated features, such as elementary features and the operations used to construct these compound aKDFE features, are identified and stored. Additional metadata are automatically generated, including: (i) the number of elementary features within a compound feature, (ii) the iteration at which the feature was generated, and (iii) statistical distributions of feature values, such as mean and standard deviation.

## 3.5 Machine learning pipeline for aKDFE validation

To quantify the impact of aKDFE, two distinct machine learning pipelines were employed: (A) a machine learning pipeline designed for patient-centric data formatted feature sets, and (B) a machine learning pipeline tailored for event-based feature sets. The primary distinction between these pipelines lies in the prediction models utilised. Pipeline (A) implemented feedforward neural networks, while pipeline (B) employed a recursive model specifically adapted for features with temporal information.

To broaden the scope of prediction model evaluation, event-based feature sets were also assessed using pipeline (A).

The schematic overview of machine learning pipeline A is illustrated in Figure 6.

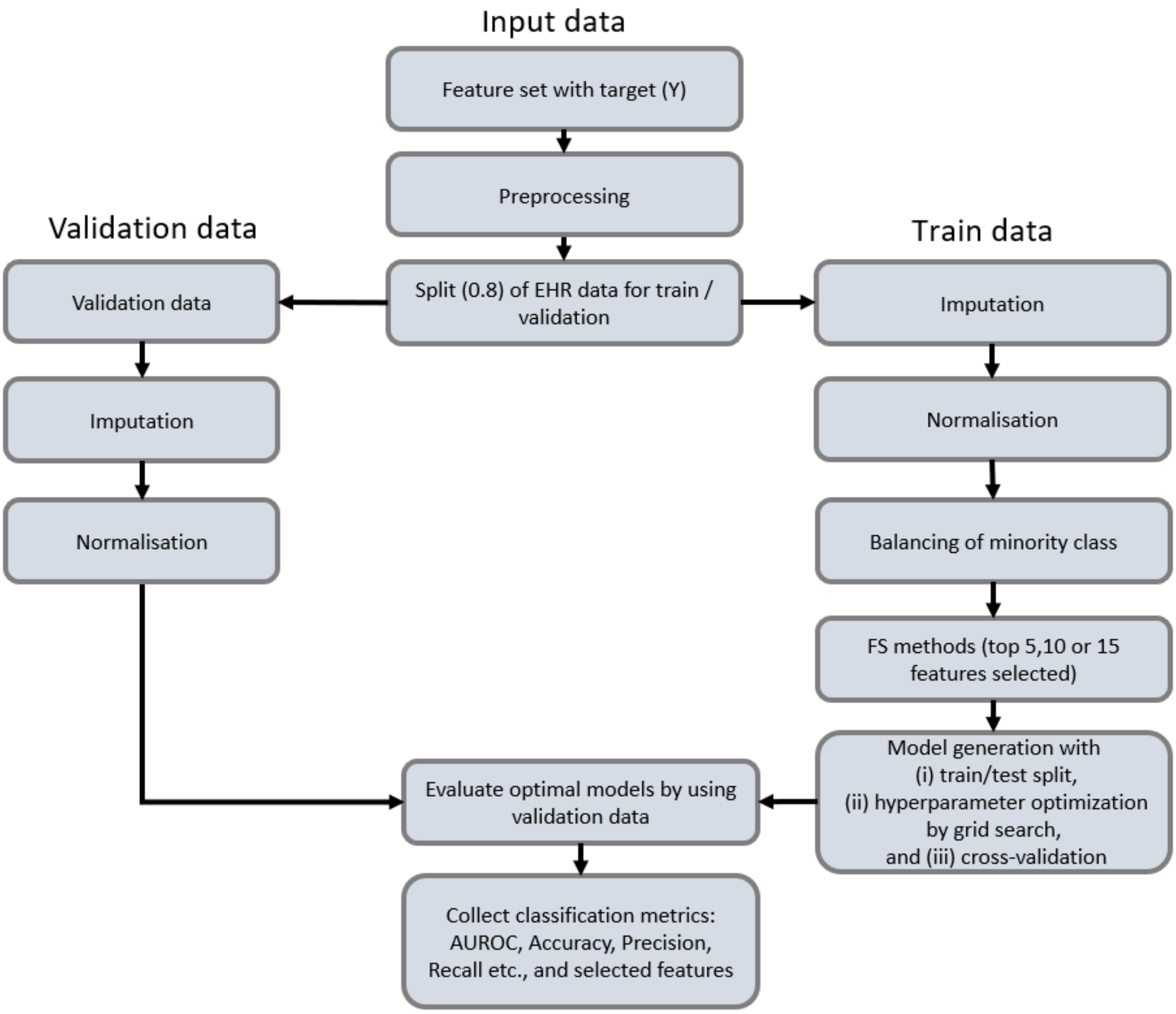

*Figure 6 Schematic overview of processes in machine learning pipeline A for prediction model evaluation.*

While the machine learning pipelines incorporate distinct processes, they share a common objective: to generate classification models that exhibit high performance and validity with minimal resource utilisation. For a comprehensive description of the machine learning pipeline stages, processes, and methodologies, cf. the Appendix.

## 3.6 Evaluation and performance metrics

The AUROC metric was implemented using the python/scikit-learn library. For detailed information regarding AUROC, cf. [60] and the scikit-learn documentation [61].

To analyse the classification performance of aKDFE, hypothesis testing was conducted. While several quantitative classification metrics were collected from the experiments, as stated in Section 2.10, AUROC is considered a salient metric and is frequently employed in similar studies involving EHR data [3, 55]. Therefore, this study compared the predictive performance of generated prediction models using AUROC. Hypothesis testing was performed using ANOVA based on AUROC values. Statistical significance was determined using a significance level of 0.05 [62].

## 3.7 Performed experiments

To evaluate the hypotheses H1 and H2 a series of experiments was conducted using the generated feature sets detailed in Table 3 and Figure 3. For the event-based feature sets, where patients can be represented by multiple rows, patients may exhibit varying predicted outcomes.

For instance, the same patient could be assigned both class 1 and class 0. Event-based feature sets contain valuable inherent information regarding the temporal order of events, which has the potential to enhance prediction models for ADEs [63]. To assess the potential of feature sets incorporating temporal information, additional experiments were performed using a recursive prediction model, see Table 4.

*Table 4 Description of performed experiments in relation to used feature sets.*

| Experiment number | Experiment name | Feature set used |
|---|---|---|
| 1 | E1-EVENT | 1-EVENT |
| 2 | E1-EVENT-aKDFE | 1-EVENT-aKDFE |
| 3 | E1-PCD-aKDFE | 1-PCD-aKDFE |
| 4 | E2-EVENT | 2-EVENT |
| 5 | E2-EVENT-aKDFE | 2-EVENT-aKDFE |
| 6 | E2-EVENT-aKDFE-RNN | 2-EVENT-aKDFE |
| 7 | E2-PCD-aKDFE | 2-PCD-aKDFE |

Based on our experience, experiment E1-EVENT represents a feature set commonly employed in standard medical registry research projects. These projects often involve manual feature engineering, similar to the characteristics observed in E1-EVENT. More advanced research projects incorporate state-of-the-art domain knowledge from validated knowledge databases or CDSSs. Experiment E2-EVENT represents these more sophisticated projects.

As illustrated in Table 5, to evaluate hypotheses H1 and H2, the variance of AUROC values for Experiment 1 and Experiment 2 was compared for each sub-hypothesis using ANOVA. For specific about sub-hypotheses and corresponding experiments, see Table 5.

*Table 5 Performed sub-hypothesis to test the hypotheses H1, and H2. PCD stands for Patient-Centric Data.*

| Hypothesis | Sub-hypothesis | Experiment 1 | Experiment 2 |
|---|---|---|---|
| H1 | H1.1 | E1-EVENT | E1-EVENT-aKDFE |
| H1 | H1.2 | E2-EVENT | E2-EVENT-aKDFE |
| H2 | H2.1 | E1-EVENT | E2-EVENT |
| H2 | H2.2 | E1-EVENT-aKDFE | E2-EVENT-aKDFE |
| H2 | H2.3 | E1-PCD-aKDFE | E2-PCD-aKDFE |
| H1 | H1.3 | E1-PCD-aKDFE | Maximum AUC of all event-based experiments |
| H1 | H1.4 | E2-PCD-aKDFE | Maximum AUC of all event-based experiments |

To evaluate hypothesis H1, the results from sub-hypotheses H1.1 through H1.4 were validated. To evaluate hypothesis H2, the results from sub-hypotheses H2.1 and H2.2 were validated. To validate both steps in the aKDFE process, the additional hypotheses H1.3 and H1.4 were evaluated.

## 3.8 Ethical considerations

This research was approved by the Regional Ethical Review Authority (*Swedish Etikprövningsmyndigheten*, dnr 2021-03880). Informed consent of participants was not required, as this retrospective study did not impact the healthcare of the included patients. Social security numbers (*Swedish personnummer*) were not disclosed during the study, instead, a pseudonymised code was utilised to represent each unique patient.

# 4 Results

This study aims to evaluate the classification performance of feature sets generated by aKDFE, specifically when metrics from a CDSS are incorporated. This approach quantifies the effectiveness of integrating CDSS metrics. The study's focus was not on fine-tuning individual FE methods, classification models, or other components of the machine learning pipeline.

The results section presents: (i) a summary of the selected features, (ii) the classification metrics for each experiment, and (iii) the results of hypothesis testing to validate differences between experiments.

## 4.1 Included prediction models in the machine learning pipelines

The vast number of potential methods available for inclusion in a machine learning pipeline, coupled with the multitude of options within each pipeline stage, necessitates a comprehensive search for an optimal prediction model, here approached from an automated machine learning (AutoML) perspective [64]. However, the extent of model generation is primarily constrained by computational resources and time.

### 4.1.1 Included models in the feedforward machine learning pipeline (A)

To address the performance challenges encountered with feedforward-models, a more comprehensive machine learning pipeline was implemented in the E1-EVENT experiment. Subsequently, based on the AUROC results, a streamlined machine learning pipeline was adopted for the remaining experiments. For detailed information regarding the methods employed within the machine learning pipeline, cf. the Appendix.

The models generated using the more extensive machine learning pipeline in the E1-EVENT experiment were evaluated. The evaluation revealed that complex and computationally intensive models, such as (i) Random Forest (RF), (ii) Multilayer Perceptron (MLP), a type of neural network, (iii) Support Vector Machine (SVM) with non-linear kernels, and (iv) Gradient Boosting (GB), exhibited prolonged execution times. This was attributed to both their inherent complexity and the extensive parameter grid search conducted for each model. Table 6 provides detailed information about the performed experiments.

*Table 6 Classification metrics for extensive machine learning pipeline used in experiment E1-EVENT.*

| Experiment | Model | Number of model evaluations | Mean AUC | Mean execution time (hours) |
|---|---|---|---|---|
| E1-EVENT | SVM | 1 | Did not finish | Did not finish |
| E1-EVENT | Random forest | 12 | 0.826 | 3.3 |
| E1-EVENT | KNN | 12 | 0.795 | 0.5 |
| E1-EVENT | Logistic regression | 12 | 0.732 | 0.03 |
| E1-EVENT | Linear SVM | 12 | 0.726 | 0.1 |
| E1-EVENT | MLP | 1 | 0.748 | 5.1 |
| E1-EVENT | Gradient boosting | 2 | 0.79 | 19.4 |
| E1-EVENT | Total | 52 | 0.77 | 1.8 |

To establish a practical machine learning pipeline for the extensive event-based feature sets—enabling both the evaluation of the aKDFE method and adherence to the project's computational resource limitations—a standardisedpipeline was employed across all experiments, see Table 7.

Additionally, the specific machine learning pipeline used for the patient-centric data formatted feature set group is also included. To assess the performance of computationally intensive models, such as RF and MLP, on the EVENT feature set, a supplementary model evaluation was conducted using optimised parameters and pipeline methods for RF and MLP, see Table 7.

*Table 7 Included and evaluated models in performed machine learning pipeline A.*

| Feature set group | Imputation | Normalisation | Balancing | Feature selection | Model (Number of results) |
|---|---|---|---|---|---|
| event EHR, event aKDFE | Mean, median | MinMaxScaler | SMOTE, ADASYN | ANOVA F-score, chi2 | KNN (8), Linear SVM (8), logistic regression (8) |
| event EHR, event-aKDFE | Mean | MinMaxScaler | ADASYN | ANOVA F-score | MLP (1) |
| event EHR, event-aKDFE | Mean | MinMaxScaler | ADASYN | ANOVA F-score | Random forest (1) |
| Patient-centric data formatted aKDFE | Mean, median | MinMaxScaler | SMOTE, ADASYN, none | ANOVA F-score, chi2, mutual information | KNN (18), SVM (18), logistic regression (18), random forest (18) |

### 4.1.2 Included models in the recursive machine learning pipeline (B)

Experiment E2-EVENT-aKDFE utilised feature set 2-EVENT-aKDFE and employed the recursive machine learning pipeline B to generate an RNN prediction model.

### 4.1.3 Summary of included prediction models

To summarise the results obtained from the prediction model generation and evaluation, 26 result sets were recorded for each experiment within the EVENT EHR feature set group, and 72 result sets for each experiment within the patient-centric data formatted aKDFE feature set group. A *baseline machine learning pipeline* was established for the EVENT feature set groups. This baseline pipeline comprised three models: KNN, linear SVM, and logistic regression, along with standard machine learning pipeline procedures. Consequently, each experiment in the EVENT group yielded 24 result sets, as the results for RF and MLP were excluded in the baseline machine learning pipeline. For the patient-centric data formatted feature set groups, a total of 72 result sets were collected, as detailed in Table 7.

## 4.2 Selected features

The machine learning pipeline A incorporated a FS process that evaluated three distinct FS methods. Model generation employed cross-validation to assess performance across three feature subset sizes: 5, 10, and 15 selected features. A comprehensive summary of the selected features, including detailed descriptions, is provided in the Appendix. To identify the most influential features, we examined the five best-performing prediction models for each experiment, as determined by the AUROC metric.

The results of the four event-based experiments are presented separately in Sections 4.2.1 (event EHR group) and 4.2.2 (event aKDFE group) due to the disparity in the number of features between these sets. Furthermore, Section 4.2.3 details the selected features for the two feature sets within the patient-centric data formatted aKDFE feature set group.

### 4.2.1 Event EHR feature set group experiments

Experiments were conducted on the event EHR feature set group, utilising two distinct feature sets: (i) 1-EVENT and (ii) 2-EVENT. The key difference between these sets lies in the inclusion of domain knowledge regarding ADR risk within 2-EVENT. This knowledge was represented as a drug risk score derived from Janusmed. Consequently, the feature sets varied in size: 1-EVENT contained 11 features, while 2-EVENT comprised 13.

Within the machine learning pipeline, categorical values were encoded into separate binary features. Specifically, feature id 3 (gender) and feature id 11 (route of administration) generated subfeatures. These subfeatures are identified by suffix letters appended to the original feature id, as detailed in Table 8.

*Table 8 Selected features for top five generated models for event EHR feature set group.*

| Performed experiment | AUROC | Selected features ids | Number of features |
|---|---|---|---|
| E1-EVENT.1 | 0.829 | 1, 2, 3a, 3b, 4, 5, 6, 7, 8, 10, 12a, 12b, 12c, 12d, 12e | 15 |
| E1-EVENT.2 | 0.829 | 1, 2, 3a, 3b, 4, 5, 6, 7, 8, 10, 12a, 12b, 12c, 12d, 12e | 15 |
| E1-EVENT.3 | 0.829 | 1, 2, 3a, 3b, 4, 5, 6, 7, 8, 10, 12a, 12b, 12c, 12d, 12e | 15 |
| E1-EVENT.4 | 0.829 | 1, 2, 3a, 3b, 4, 5, 6, 7, 8, 10, 12a, 12b, 12c, 12d, 12e | 15 |
| E1-EVENT.5 | 0.829 | 1, 2, 3a, 3b, 4, 5, 6, 7, 8, 10, 12a, 12b, 12c, 12d, 12e | 15 |
| E2-EVENT.1 | 0.829 | 1, 2, 3a, 3b, 4, 5, 6, 7, 8, 10, 12a, 12b, 12c, 12d, 12e | 15 |
| E2-EVENT.2 | 0.815 | 1, 2, 3a, 3b, 4, 5, 6, 7, 8, 10, 12a, 12b, 12c, 12d, 12e | 15 |
| E2-EVENT.3 | 0.814 | 1, 2, 3a, 3b, 4, 5, 6, 7, 8, 10, 12a, 12b, 12c, 12d, 12e | 15 |
| E2-EVENT.4 | 0.814 | 1, 2, 3a, 3b, 4, 5, 6, 7, 8, 10, 12a, 12b, 12c, 12d, 12e | 15 |
| E2-EVENT.5 | 0.813 | 1, 2, 3a, 3b, 4, 5, 6, 7, 8, 10, 12a, 12b, 12c, 12d, 12e | 15 |

For the corresponding features names to the feature ids in Table 8, see Table 9.

*Table 9* Feature names and ids for selected features for top five AUROC scores for prediction models in the event EHR feature set group.

| Feature name | Feature id | Feature description |
|---|---|---|
| PATIENT_STUDY AGE DECADE | 1 | Patient age decade at start of study |
| EVENT CONCEPT TYPE ID | 2 | Type of event |
| GENDER-F | 3a | Patient gender (female) |
| GENDER-M | 3b | Patient gender (male) |
| PATIENT AGE AT OBSERVATION | 4 | Patient age at event |
| CENSOR_DATE | 5 | Date of patient death or censoring |
| DRUG USE INDEX GROUP | 6 | Estimated index of patient drug use |
| OBSERVATION_START_DATE | 7 | Date for event |
| VALUE CHAR | 8 | ATC Code or ICD-10 diagnosis code |
| DRUG SUBSTANS ID | 11 | Drug substance id |
| ROUTE OF ADMINISTRATION - missing | 12a | Route of administration type (missing) |
| ROUTE OF ADMINISTRATION TYPE-OLS | 12b | Route of administration type (OLS - oral floating, or semi firm) |
| ROUTE OF ADMINISTRATION TYPE-OSD | 12c | Route of administration type (OSD - oral firm) |
| ROUTE OF ADMINISTRATION TYPE-PAR | 12d | Route of administration type (PAR - parenteral) |
| ROUTE OF ADMINISTRATION TYPE-REC | 12e | Route of administration type (REC - rectal) |

### 4.2.2 Event aKDFE feature set group experiments

Experiments using the event aKDFE feature group were conducted on two feature sets: (i) 1-EVENT-aKDFE and (ii) 2-EVENT-aKDFE. The 2-EVENT-aKDFE set incorporates domain knowledge regarding ADR risk, specifically a drug risk score derived from Janusmed, see Table 10.

*Table 10 Selected features for top five generated models for event aKDFE feature set group.*

| Performed experiment | AUROC | Selected features ids | Number of features |
|---|---|---|---|
| E1-EVENT-aKDFE.1 | 0.917 | 3a, 3b, 4, 5, 7 | 5 |
| E1-EVENT-aKDFE.2 | 0.912 | 2, 3a, 3b, 4, 5, 7, 8, 9 | 8 |
| E1-EVENT-aKDFE.3 | 0.911 | 2, 3a, 3b, 4, 5, 7, 8, 9 | 8 |
| E1-EVENT-aKDFE.4 | 0.911 | 2, 3a, 3b, 4, 5, 7, 8, 9 | 8 |
| E1-EVENT-aKDFE.5 | 0.899 | 2, 3a, 3b, 4, 5, 7, 8, 9 | 8 |
| E2-EVENT-aKDFE.1 | 0.921 | 3a, 3b, 4, 5, 7 | 5 |
| E2-EVENT-aKDFE.2 | 0.909 | 2, 3a, 3b, 4, 5, 7, 8, 9 | 8 |
| E2-EVENT-aKDFE.3 | 0.908 | 2, 3a, 3b, 4, 5, 7, 8, 9 | 8 |
| E2-EVENT-aKDFE.4 | 0.908 | 2, 3a, 3b, 4, 5, 7, 8, 9 | 8 |
| E2-EVENT-aKDFE.5 | 0.902 | 3a, 3b, 4, 5, 7 | 5 |

For the corresponding features names to the feature ids in Table 10, cf. Table 11.

*Table 11 Feature names and ids for selected features for top five AUROC scores for prediction models in the event aKDFE feature set group.*

| Feature name | Feature id | Feature description |
|---|---|---|
| FEATURE ID | 2 | Feature id for aKDFE generated feature |
| GENDER | 3 | Patient gender |
| PATIENT AGE AT OBSERVATION | 4 | Patient age at feature id event |
| CENSOR_DATE | 5 | Date of patient death or censoring |
| OBSERVATION START DATE | 7 | Date for feature id event |
| VALUE_CHAR | 8 | Character value for Feature id |
| VALUE DECIMAL | 9 | Decimal value |

### 4.2.3 Patient-centric data formatted aKDFE feature set group experiments

Experiments were conducted on the patient-centric data formatted aKDFE feature group, utilising two distinct feature sets: (i) 1-PCD-aKDFE and (ii) 2-PCD-aKDFE. The key difference between these sets is that 2-PCD-aKDFE incorporates domain knowledge regarding ADR risk from Janusmed, represented as a drug risk score. As detailed in Section 3.3.2, two different patient-centric transformation methods were employed. Each method resulted in features with unique prefixes: (i) the number of N-grams (FC, representing Feature Count) and (ii) the sum of feature values (S, representing Sum), as outlined in Table 12.

*Table 12 Description of patient-centric data formatted aKDFE feature set groups.*

| Patient-centric data formatted feature set | Type | Naming prefix | N | Feature description |
|---|---|---|---|---|
| 1-PCD-aKDFE | N-gram | FC | 528 | Transformed features (N-grams) |
| 1-PCD-aKDFE | Sum | S | 883 | Transformed features (sum feature value) |
| 2-PCD-aKDFE | N-gram | FC | 486 | Transformed features (N-grams) |
| 2-PCD-aKDFE | Sum | S | 843 | Transformed features (sum feature value) |

A summary of the selected features for the patient-centric data formatted aKDFE feature sets, as identified by the five best prediction models (based on AUROC) for each experiment, is presented in Table 13.

*Table 13 Selected features for top five generated models for patient-centric data formatted aKDFE feature set group based on AUROC.*

| Performed experiment | AUROC | Selected features ids | Number of features |
|---|---|---|---|
| E1-PCD-aKDFE.1 | 0.998 | FC1000, FC1044, FC1352, FC1560, FC1616, FC1619, FC2029, FC2509, FC2510, FC2699, FC2961, FC545, FC999, S2699, S5608 | 15 |
| E1-PCD-aKDFE.2 | 0.998 | FC1000, FC1044, FC1352, FC1560, FC1616, FC1619, FC2029, FC2509, FC2510, FC2699, FC2961, FC545, FC999, S2699, S7525 | 15 |
| E1-PCD-aKDFE.3 | 0.997 | FC1352, FC1620, FC2221, FC2445, FC2509, FC2510, FC2660, FC2699, FC2719, FC2949, FC2950, FC2961, FC3015, FC999, S3809 | 15 |
| E1-PCD-aKDFE.4 | 0.997 | FC1352, FC1620, FC2221, FC2445, FC2509, FC2510, FC2660, FC2699, FC2719, FC2949, FC2950, FC2961, FC3015, FC999, S3809 | 15 |
| E1-PCD-aKDFE.5 | 0.997 | FC1352, FC1620, FC2221, FC2445, FC2509, FC2510, FC2660, FC2699, FC2719, FC2949, FC2950, FC2961, FC3015, FC999, S3809 | 15 |
| E2-PCD-aKDFE.1 | 0.997 | FC1048, FC1049, FC1191, FC1249, FC2127, FC2727, FC2827, FC2828, FC2987, FC740, FC772, FC845, FC897, S2828, S5245 | 15 |
| E2-PCD-aKDFE.2 | 0.997 | FC1048, FC1049, FC1191, FC1249, FC2127, FC2727, FC2827, FC2828, FC2987, FC740, FC772, FC845, FC897, S11008, S2828 | 15 |
| E2-PCD-aKDFE.3 | 0.996 | FC1048, FC1191, FC1249, FC2727, FC2827, FC2828, FC2987, FC740, FC845, S3112, S3196, S6178, S6186, S6202, S7768 | 15 |
| E2-PCD-aKDFE.4 | 0.996 | FC1048, FC1191, FC1249, FC2727, FC2827, FC2828, FC2987, FC740, FC845, S3112, S3196, S6178, S6186, S6202, S7768 | 15 |
| E2-PCD-aKDFE.5 | 0.995 | FC1048, FC1191, FC1249, FC740, FC845 | 5 |

### 4.2.4 Correlation between selected features and target class for patient-centric data formatted feature sets

To examine the relationship between the selected features, presented in Table 13, and the target class, Pearson correlation matrices were generated. Specifically, we present: (i) Figure 7 reporting the five features with the highest individual correlation for feature set for 1-PCD-aKDFE, and (ii) Figure 8 for feature set 2-PCD-aKDFE.

Secondly the 15 selected features for the experiments with the highest AUROC for the experiments, (iii) E1-PCD-aKDFE Figure 9, and (iv) E2-PCD-aKDFE Figure 10.

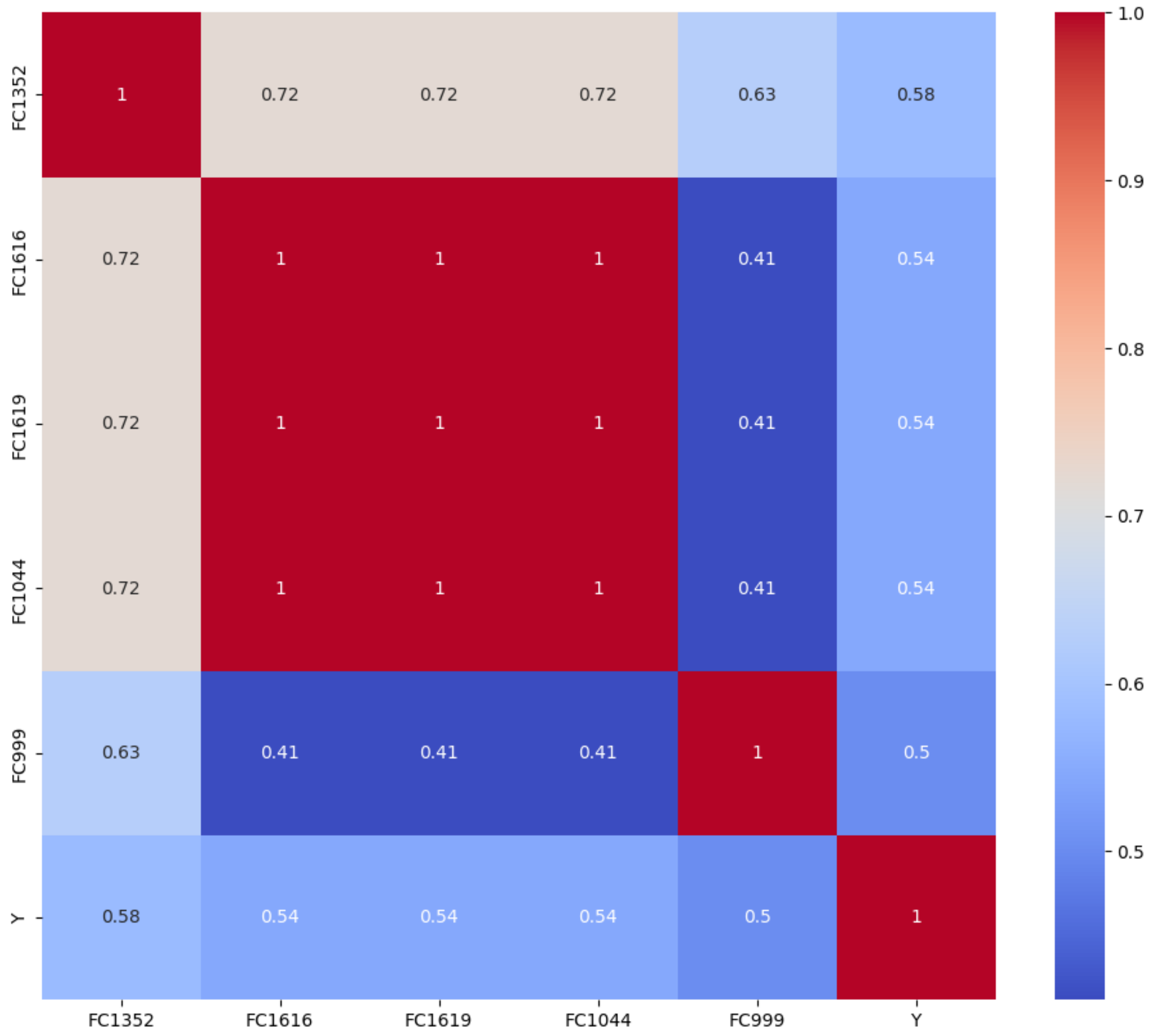

*Figure 7 Correlation matrix based on Pearson coefficient for the five features with the highest correlation for feature set 1-PCD-aKDFE.*

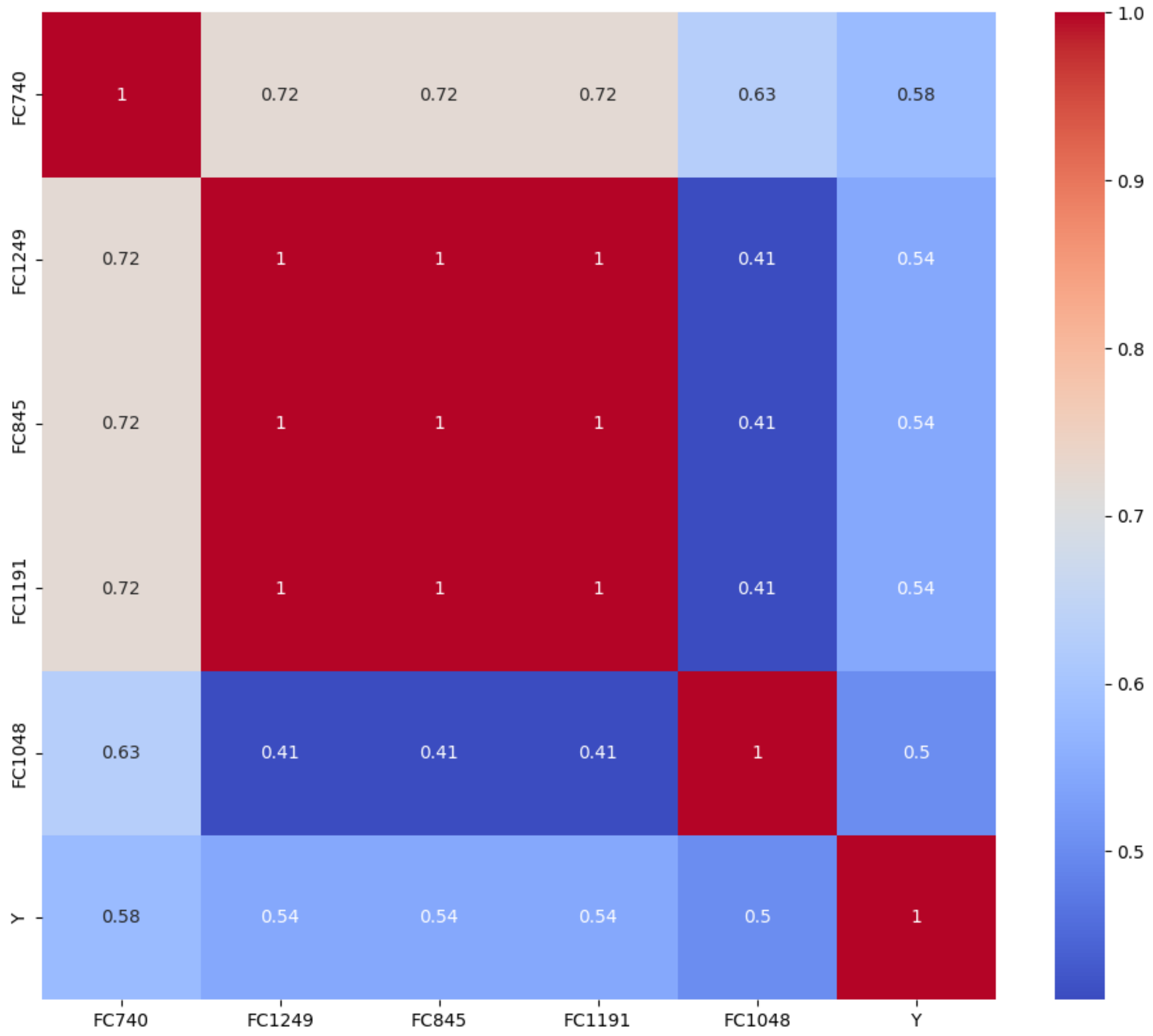

*Figure 8 Correlation matrix based on Pearson coefficient for the five features with the highest correlation for feature set 2-PCD-aKDFE.*

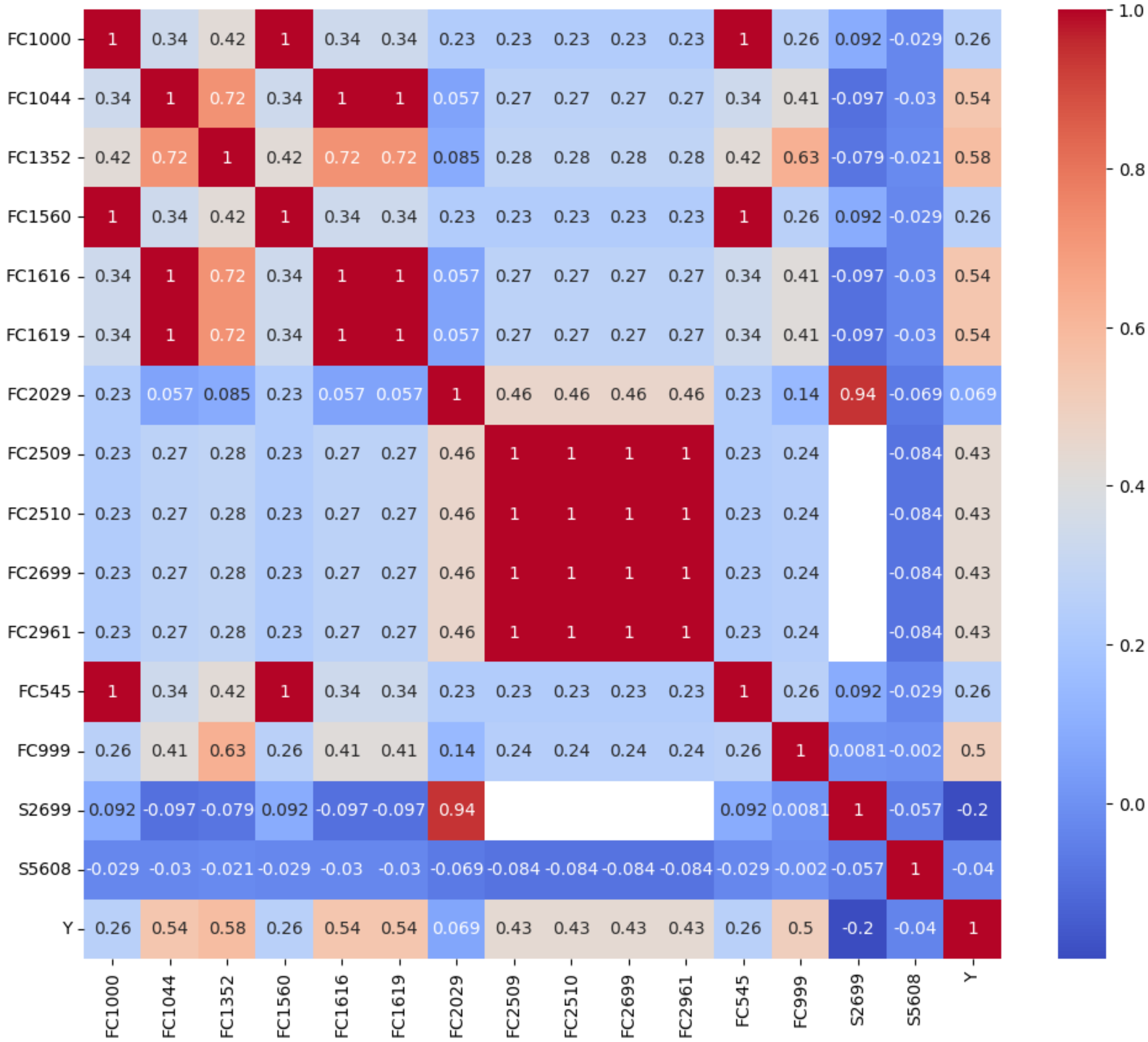

*Figure 9 Correlation matrix based on Pearson coefficient for the 15 selected features in experiment E1-PCD-aKDFE.1.*

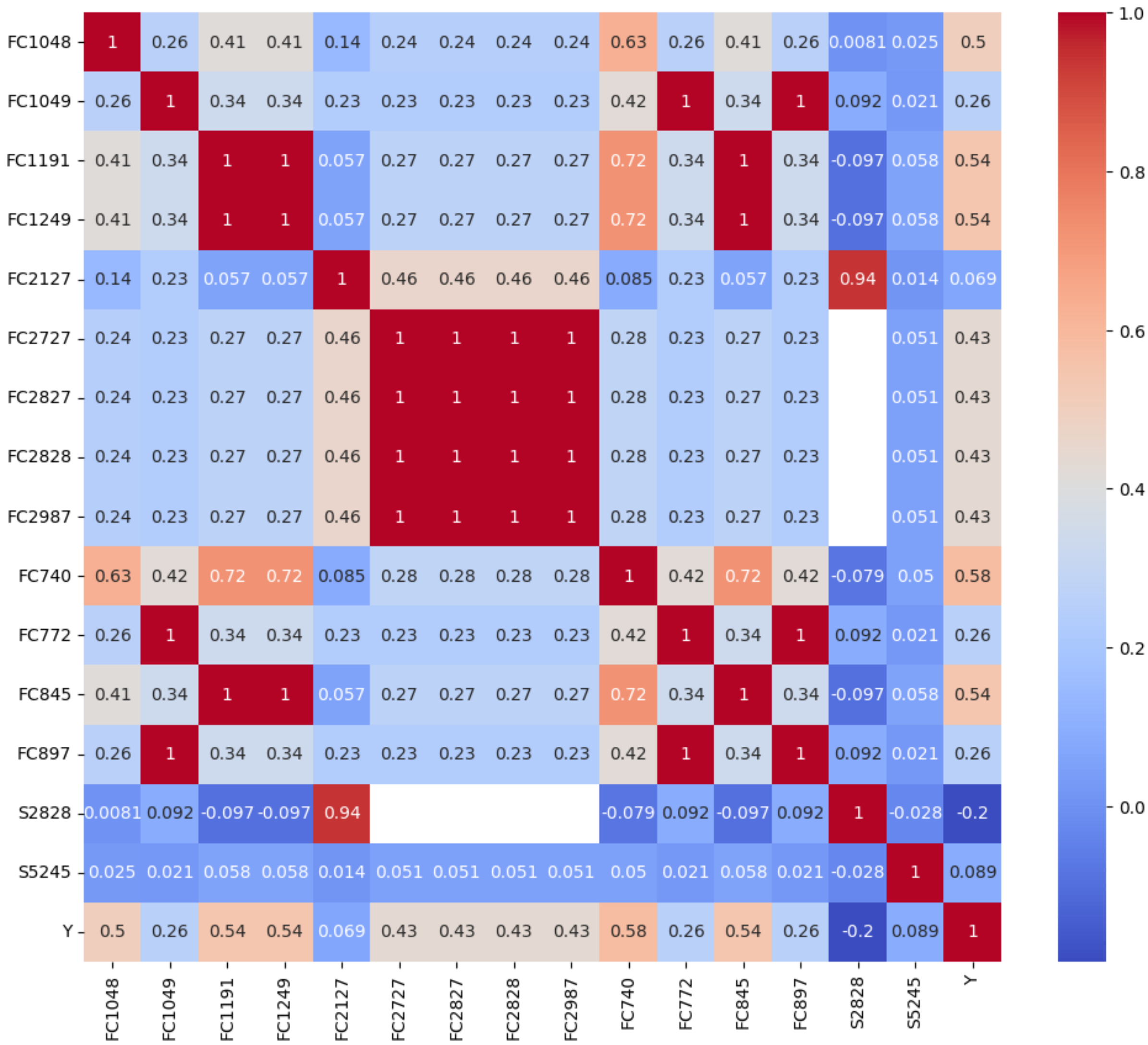

*Figure 10 Correlation matrix based on Pearson coefficient for the 15 selected features in experiment E2-PCD-aKDFE.1.*

Table 14 details the three features with the highest correlation for each patient-centric data formatted feature set.

*Table 14 Description of the three features with the highest correlation to the target variable for the patient-centric data formatted aKDFE feature set group. The feature ids are unique for each feature set, but the feature code is the same.*

| Feature set | Feature id | Feature code | Correlation value | Description |
|---|---|---|---|---|
| 1-PCD-aKDFE | F1352 | H0_2065=I49 | 0.58 | H0 = First parent level for the diagnosis code, 2065 = Diagnosis for ventricular arrythmia. The patient has a diagnosis code for ventricular arrythmia belonging to the diagnosis category I49. |
| 1-PCD-aKDFE | FC1616 | 2065=I499 | 0.54 | 2065 = Diagnosis for ventricular arrythmia. The patient has the diagnosis code for ventricular arrythmia I499. |
| 1-PCD-aKDFE | FC1619 | 2100=I499 | 0.54 | 2100 = Diagnosis for arrythmia. The patient has the diagnosis code for arrythmia I499. |
| 2-PCD-aKDFE | FC740 | H0_2065=I49 | 0.58 | H0 = First parent level for the diagnosis code, 2065 = Diagnosis for ventricular arrythmia. The patient has a diagnosis code for ventricular arrythmia belonging to the diagnosis category I49. |
| 2-PCD-aKDFE | FC1249 | 2065=I499 | 0.54 | 2065 = Diagnosis for ventricular arrythmia. The patient has the diagnosis code for ventricular arrythmia I499. |
| 2-PCD-aKDFE | FC845 | 2100=I499 | 0.54 | 2100 = Diagnosis for arrythmia. The patient has the diagnosis code for arrythmia I499. |

### 4.2.5 Correlation between features and target class for CDSS Janusmed features

To evaluate the relevance of CDSS Janusmed related features based on prediction model generation only the features that include Janusmed information are reported in Figure 11 for patient-centric transformed N-gram features and in Figure 12 for patient-centric transformed sum features values. For a more detailed description, see Table 15.

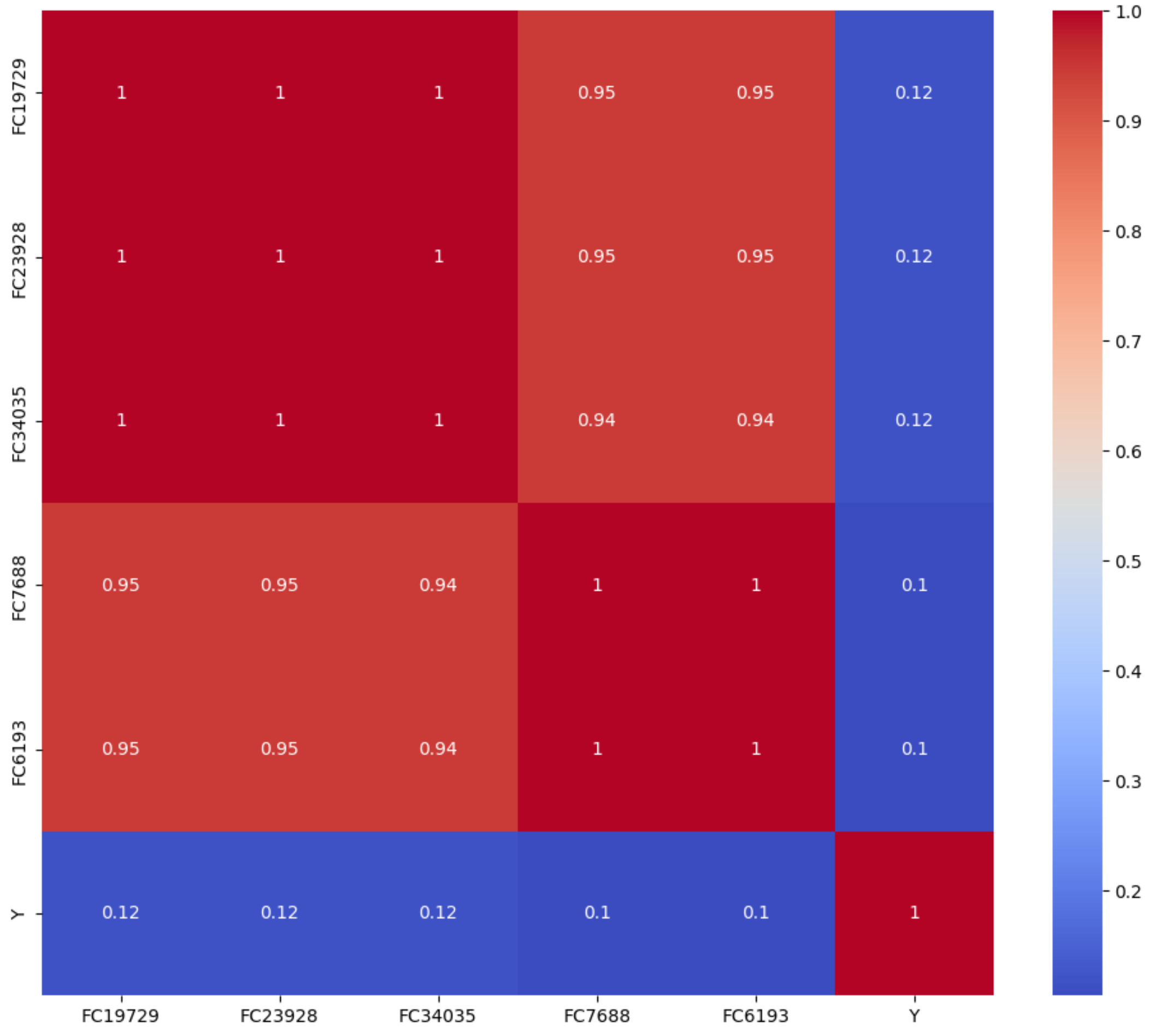

*Figure 11 Correlation matrix based on Pearson coefficient for the five features with the highest correlation for features related to CDSS Janusmed and patient-centric transformation N-grams in the feature set 2-PCD-aKDFE.*

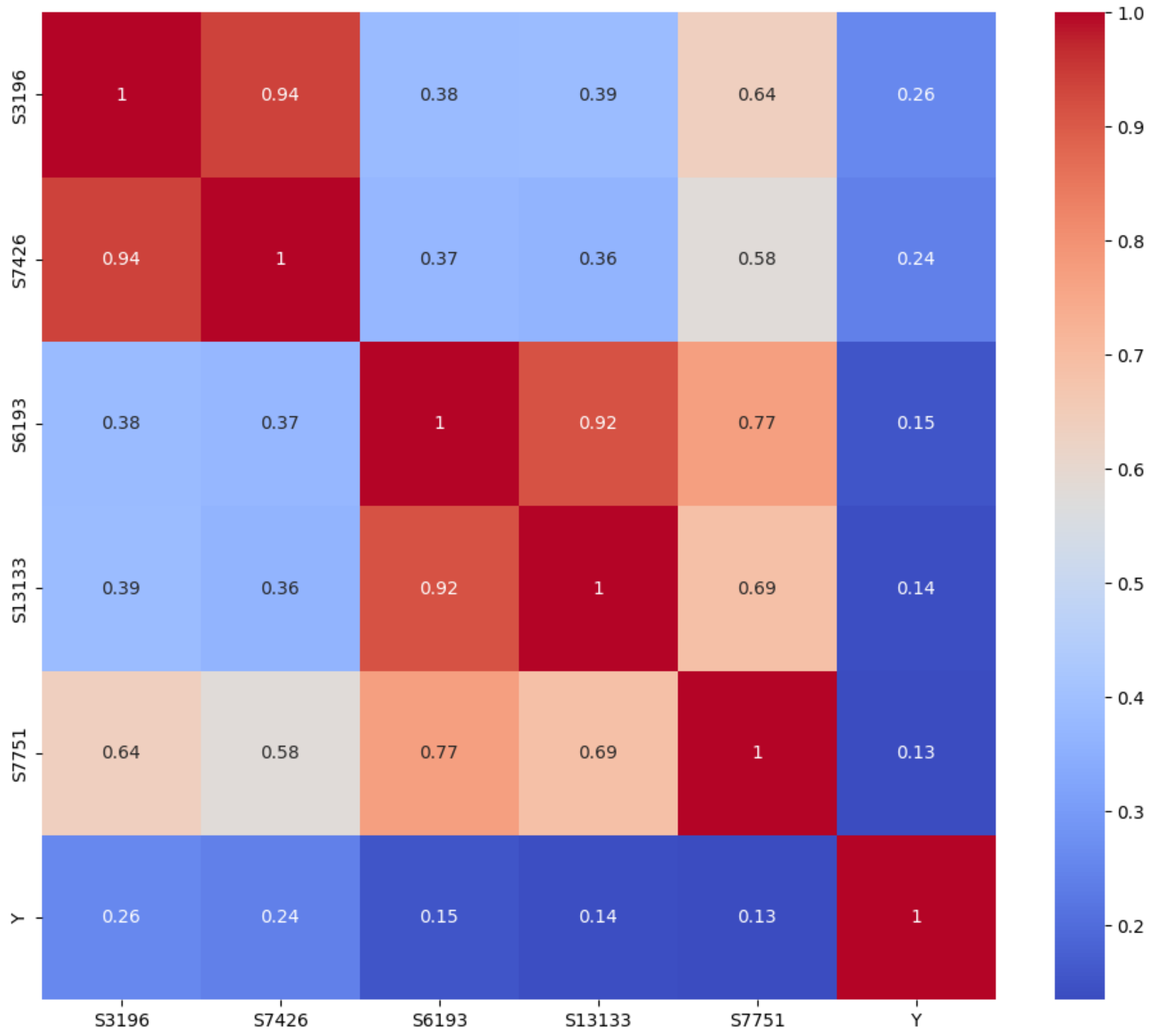

*Figure 12 Correlation matrix based on Pearson coefficient for the five features with the highest correlation for features related to CDSS Janusmed and patient-centric transformation sum value in the feature set 2-PCD-aKDFE.*

In summary, Table 15 describes the top two features for the patient-centric transformations N-gram and sum feature value that originated from the CDSS Janusmed feature set.

*Table 15 Description of the best two features, for the patient-centric transformations N-gram and sum feature value, with the highest correlation to the target variable for the features in 2-PCD aKDFE that are related to CDSS Janusmed. The feature ids are unique for each feature set, but the feature code is the same.*

| Feature set | Feature id | Feature code | Correlation value | Description |
|---|---|---|---|---|
| 2-PCD-aKDFE | FC19729 | 2105=ALL-1060-2007=DF_JM_1-1040-1033 | 0.12 | 2105 = All diagnosis, 1060 = Last event of many, 2007 = Route of administration, 1040 = Have observation (1/0), 1033 = Days before a date.<br>The number of days from the last of all diagnosis events to a drug event for an enteral (oral) drug. |
| 2-PCD-aKDFE | FC23928 | 2105=ALL-1060-2007=DF_JM_1-1070-1033 | 0.12 | 1070 = Number of observations.<br>The number of days from the last of all diagnosis events to the date for the |
| | | | | calculation of the number of drug events for enteral (oral) drugs. |
| 2-PCD-aKDFE | S3196 | 2007=DF_JM_1-1060-2105=ALL-1050-1030 | 0.26 | 1030 = Days between, 1050 = First event of many The sum of all days between the last drug event for enteral (oral) drugs and the first diagnosis event of many. |
| 2-PCD-aKDFE | S7426 | 2006=RL_1,000-1060-2105=ALL-1050-1030 | 0.24 | 2006 = Janusmed risk level. The number of days between the last Janusmed risk level = 1 and the first diagnosis event of all diagnosis. |

## 4.3 Classification results

To compare the predictive performance of the feature set groups, the outcome Y was predicted using the machine learning classification pipeline and evaluated with various metrics, cf. Section 3.6.

### 4.3.1 Classification metric

Table 16 summarises the classification performance of the top three prediction models, selected based on AUROC, for each experiment within the baseline machine learning pipeline. In addition to AUROC, other relevant performance metrics are reported. These results correspond to the models detailed in Section 4.1.

*Table 16 Mean prediction metrics from generated models from the baseline machine learning pipeline (A) and recursive pipeline (B).*

| Experiment | Model | Accuracy | Precision | Recall | F1-score | AUROC | Log loss | Brier score |
|---|---|---|---|---|---|---|---|---|
| E1-EVENT | KNN | 0.85 | 0.888 | 0.849 | 0.862 | 0.811 | 5.444 | 0.151 |
| E1-EVENT | KNN | 0.849 | 0.888 | 0.849 | 0.862 | 0.811 | 5.444 | 0.151 |
| E1-EVENT | KNN | 0.849 | 0.888 | 0.849 | 0.862 | 0.811 | 5.453 | 0.151 |
| E2-EVENT | KNN | 0.85 | 0.890 | 0.85 | 0.863 | 0.815 | 5.401 | 0.150 |
| E2-EVENT | KNN | 0.85 | 0.889 | 0.85 | 0.863 | 0.814 | 5.414 | 0.150 |
| E2-EVENT | KNN | 0.85 | 0.889 | 0.849 | 0.862 | 0.814 | 5.458 | 0.151 |
| E1-EVENT-aKDFE | KNN | 0.977 | 0.977 | 0.977 | 0.977 | 0.917 | 0.824 | 0.023 |
| E1-EVENT-aKDFE | KNN | 0.932 | 0.954 | 0.932 | 0.939 | 0.912 | 2.467 | 0.068 |
| E1-EVENT-aKDFE | KNN | 0.949 | 0.959 | 0.949 | 0.952 | 0.911 | 1.854 | 0.051 |
| E2-EVENT-aKDFE | KNN | 0.979 | 0.979 | 0.979 | 0.979 | 0.921 | 0.763 | 0.021 |
| E2-EVENT-aKDFE | KNN | 0.930 | 0.953 | 0.93 | 0.938 | 0.909 | 2.514 | 0.070 |
| E2-EVENT-aKDFE | KNN | 0.948 | 0.958 | 0.948 | 0.951 | 0.908 | 1.881 | 0.052 |
| E1-PCD aKDFE | Random forest | 0.996 | 0.996 | 0.996 | 0.996 | 0.998 | 0.156 | 0.004 |
| E1-PCD aKDFE | SVM | 0.996 | 0.996 | 0.996 | 0.996 | 0.998 | 0.156 | 0.004 |
| E1-PCD aKDFE | SVM | 0.995 | 0.995 | 0.995 | 0.995 | 0.997 | 0.180 | 0.005 |
| E2-PCD aKDFE | Random forest | 0.995 | 0.995 | 0.995 | 0.995 | 0.997 | 0.192 | 0.005 |
| E2-PCD aKDFE | Random forest | 0.995 | 0.995 | 0.995 | 0.995 | 0.997 | 0.192 | 0.005 |
| E2-PCD aKDFE | Random forest | 0.993 | 0.993 | 0.993 | 0.993 | 0.996 | 0.264 | 0.007 |
| E2-EVENT-aKDFE-RNN | RNN | 0.504 | 0.102 | 0.200 | 0.135 | 0.486 | 2.158 | 0.411 |

Table 17 presents the classification scores for the RF models, utilising the optimal parameters and the established machine learning pipeline methods outlined in Section 4.1. A separate RF model was generated for each experiment; consequently, the results are presented independently from the baseline machine learning pipeline results shown in Table 16.

For RNN models, an initial model was developed and evaluated in experiment E2-EVENT-aKDFE-RNN. This experiment yielded a significantly low AUROC of 0.486, coupled with a substantial computational cost of 4,320 minutes on a medium-sized computer cluster. Therefore, only one RNN experiment was conducted.

*Table 17 Prediction metrics from generated random forest models with best parameters and models based on experiment E1-EVENT.*

| Experiment | Model | Accuracy | Precision | Recall | F1-score | AUROC | Log loss | Brier score |
|---|---|---|---|---|---|---|---|---|
| 1-EVENT | Random forest | 0.900 | 0.908 | 0.900 | 0.903 | 0.829 | 3.607 | 0.100 |
| 2-EVENT | Random forest | 0.897 | 0.907 | 0.897 | 0.901 | 0.829 | 3.701 | 0.103 |
| 1-EVENT-aKDFE | Random forest | 0.950 | 0.956 | 0.950 | 0.952 | 0.879 | 1.786 | 0.050 |
| 2-EVENT-aKDFE | Random forest | 0.950 | 0.955 | 0.950 | 0.952 | 0.879 | 1.817 | 0.050 |

The other predictions metrics followed the AUROC scores in terms of consistency in relation to the scoring values.

### 4.3.2 Summary of classification metrics

Figure 13 illustrates the baseline machine learning pipeline's AUROC, displaying the (i) mean, (ii) median, (iii) maximum, and (iv) RF model scores. Other classification metrics were not thoroughly assessed.

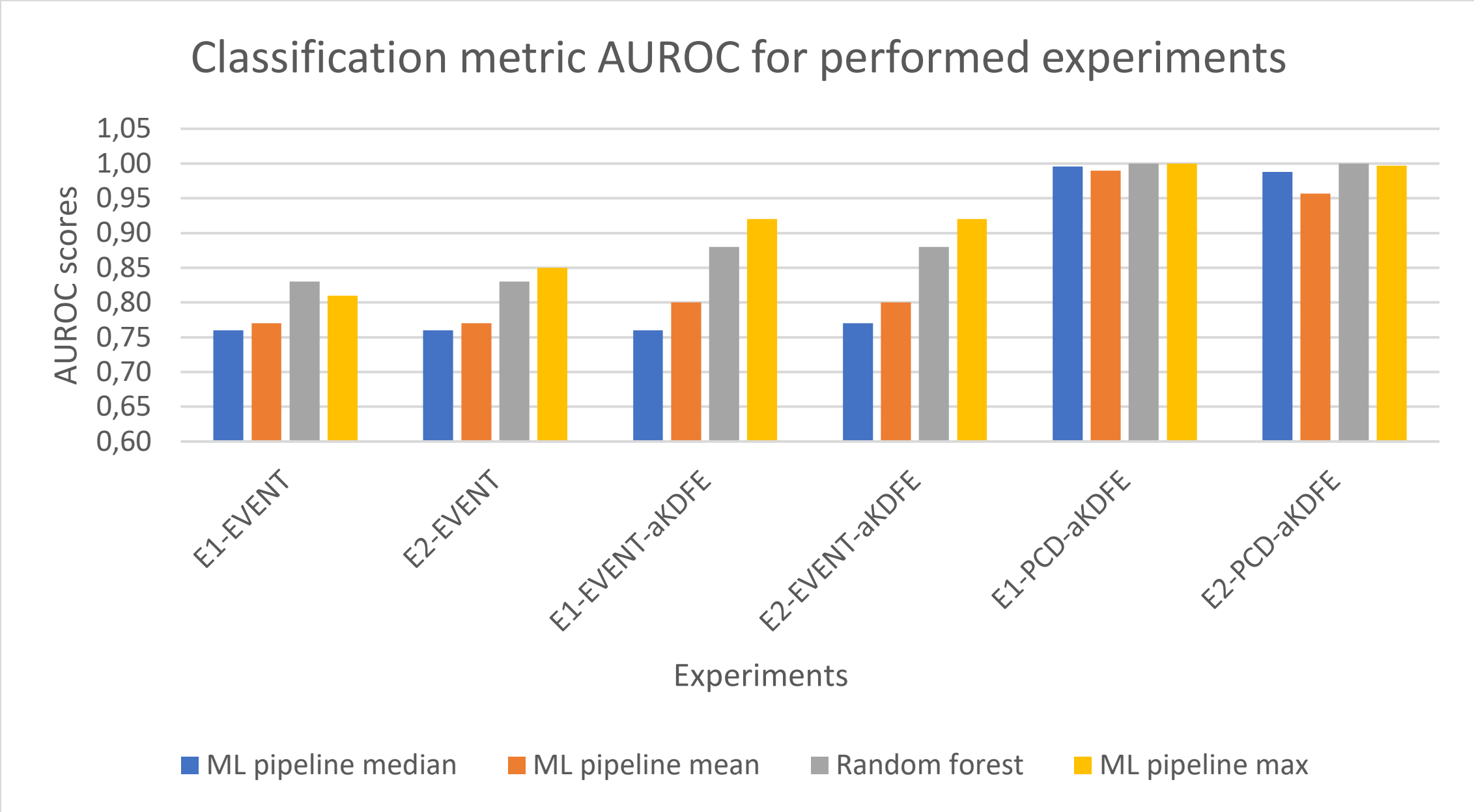

*Figure 13 Classification metric AUROC for performed experiments.*

## 4.4 Statistical analysis

To address the research questions—RQ1: "Does aKDFE generate more effective ADE prediction models compared to automated event-based KDD with incorporated domain knowledge?" and RQ2: "Does incorporating domain-specific ADE risk scores from a CDSS improve the predictive performance of ADE prediction models?"—the hypotheses presented in Table 5 were tested using ANOVA F-tests. For the event-based feature sets, the results from the baseline machine learning pipeline, specifically 24 AUROC scores per experiment, were utilised. For the patient-centric data formatted aKDFE feature sets, 72 AUROC scores were used. These results are detailed in Table 18.

*Table 18 Results from hypotheses testing based on AUROC scores.*

| Sub-hypothesis | Experiment 1 | Experiment 2 | F-score | p-value |
|---|---|---|---|---|
| H1.1 | E1-EVENT | E1-EVENT-aKDFE | 0.037 | 0.85 |
| H1.2 | E2-EVENT | E2-EVENT-aKDFE | 0.048 | 0.83 |
| H2.1 | 1-EVENT | 2-EVENT | 0.00003 | 1.00 |
| H2.2 | E1-EVENT-aKDFE | E2-EVENT-aKDFE | 0.00041 | 0.98 |
| H2.3 | E1-PCD-aKDFE | E2-PCD-aKDFE | 0.066 | 0.80 |
| H1.3 | E1-PCD-aKDFE | Maximum AUROC of all event-based experiments | 116 | < 0.00001 |
| H1.4 | E2-PCD-aKDFE | Maximum AUROC of all event-based experiments | 114 | < 0.00001 |

# 5 Discussion

This section presents the primary findings of this study, directly addressing research questions RQ1 and RQ2. Additionally, we discuss the study's aim and analyse potential threats to validity.

## 5.1 Summary of main finding

As detailed in Section 3.2.2, machine learning algorithms generally perform optimally with data in a patient-centric data format. This study provides strong statistical evidence that the transformation from event-based data to patient-centric data format, facilitated by step two of aKDFE, results in prediction models with high predictive performance, cf. Section 5.7.1. However, the inclusion of CDSS

Janusmed risk scores in the generated prediction models did not yield statistically significant improvements in prediction performance. Cf. Section 5.7.1.

## 5.2 Interpretation of results

This section presents an analysis and discussion of specific findings and reflections derived from the research.

## 5.3 High AUROC values

Several of the generated models exhibited exceptionally high AUROC values. This observation raises concerns regarding potential data leakage, where information from the output variable inadvertently influences the input features. The figures Figure 7 and Figure 9 report the correlation matrices, which illustrate strong associations between certain features and the outcome variable. While the correlation analysis in Section 4.2.4 confirms these relationships, it is crucial to acknowledge that some of these correlations are clinically plausible and verifiable within routine healthcare practices, similar finding is reported from [65]. For instance, a patient's prior healthcare history is often a significant predictor of future QT interval prolongation adverse drug events (ADEs), a finding consistent with our results.

## 5.4 Design of feed-forward machine learning pipeline (A)

The machine learning pipeline, including the selection of processes and methods for each stage, can be constructed in numerous ways. Had the study's primary aim been to maximise efficiency, specifically achieving the highest possible AUROC, a more comprehensive focus on pipeline design would have been warranted. Incorporating a greater variety of methods, conducting a wider grid search, and employing more sophisticated models would likely have yielded improved AUROC scores, albeit at a substantially increased computational cost.

To balance the objectives of (i) developing valid and high-performing prediction models and (ii) maintaining reasonable computational resource utilisation, the E1-EVENT experiment was conducted to explore a broader range of machine learning pipeline configurations. The optimal RF pipeline identified from the E1-EVENT results was subsequently applied to other experiments to estimate the highest achievable metric scores.

## 5.5 Design of recursive machine learning pipeline (B)

To leverage temporal dependencies within the medication events data for evaluating temporal relationships, a recursive machine learning model was developed and assessed in experiment E2-EVENT-aKDFE-RNN. However, the cost-benefit analysis indicated a low return, leading to the discontinuation of further recursive model development. Despite their theoretical suitability for such tasks, our results, consistent with previous findings [45], showed that recursive machine learning models did not outperform classical feed-forward machine learning models in prediction performance.

## 5.6 Entity-centric transformations

To select the appropriate transformation strategy from the alternatives outlined in Section 2.7, the selection ratio was used to evaluate the features generated in the patient-centric data formatted feature sets. The top five performing feature sets from the patient-centric data formatted aKDFE group are summarised in Table 19.

*Table 19 Number of selected features for the two patient-centric transformation methods for the five best generated prediction models for each patient-centric data formatted aKDFE feature sets, evaluated by AUROC.*

| Feature set | Total selected features | Selected patient-centric transformation method - N-gram (ratio of total) | Selected patient-centric transformation method - sum features values (ratio of total) |
|---|---|---|---|
| 1-PCD-aKDFE | 75 | 68 (0.91) | 7 (0.09) |
| 2-PCD-aKDFE | 65 | 49 (0.75) | 16 (0.25) |

The features of the N-gram patient-centric transformation method were selected more frequently than features derived from the patient-centric transformation method sum of feature values.

Figure 7 and Figure 9 present correlation matrices that illustrate the importance of each feature in relation to the target variable. Notably, some features exhibiting high inter-correlation were still selected by the machine learning pipeline. This outcome can be attributed to the FS methods employed, which were not sufficiently sophisticated to eliminate highly correlated features, a capability typically found in more advanced FS techniques. Examples of such methods include: (i) Recursive Feature Elimination (RFE), (ii) tree-based methods (e.g., Random Forest, Gradient Boosting), and (iii) Variance Inflation Factor (VIF). However, these methods demand significantly greater computational resources or processing time.

Section 4.2.4 and Table 14 demonstrate that features most strongly correlated with the target class variables were those related to the patient's pre-ADE health status, particularly concerning cardiac conditions. This finding is clinically relevant, as patients with pre-existing heart conditions are known to have an elevated risk of ADEs, specifically QT prolongation.

### 5.6.1 Baseline medical registry research

In our historical experience, a standard medical registry research project, such as those represented by E1-EVENT or E1-EVENT-aKDFE, typically relies on daily healthcare registry data. However, this mKDFE, while improving prediction models, is time-consuming and requires scarce human resources, including domain experts, data analysts, and statisticians [3].

More sophisticated research projects, exemplified by E2-EVENT or E2-EVENT-aKDFE, incorporate state-of-the-art domain knowledge from a CDSS. Theoretically, these projects should yield prediction models with superior prediction performance compared to mKDFE projects due to the integration of specialised domain knowledge.

## 5.7 Evaluation of hypotheses

### 5.7.1 Test of hypothesis H1

H1: aKDFE will generate prediction models with higher classification performance in detecting ADEs from EHR data, compared to automated event-based KDD with incorporated domain knowledge.

To evaluate hypotheses H1 and H2, we employed a significance level of 0.05, as determined by the ANOVA results presented in Section 4. Specifically, sub-hypotheses H1.1 and H1.2 could not be rejected, indicating no statistically significant evidence that aKDFE resulted in prediction models with higher AUROC values. However, sub-hypotheses H1.3 and H1.4 were rejected with strong statistical significance ($p < 0.05$). This suggests that patient-centric transformation is a highly effective technique for generating prediction models, particularly in terms of increasing AUROC.

### 5.7.2 Test of hypothesis H2

H2: Prediction models incorporating ADE risk scores from a CDSS will demonstrate improved classification performance in detecting ADEs compared to models trained without such domain expert knowledge.

To test H2, sub-hypotheses H2.1-3 were evaluated. None were rejected due to high p-values, indicating no statistically significant evidence that incorporating the CDSS Janusmed risk score into prediction models improves predictive performance.

## 5.8 Addressing the research questions

RQ1: Does aKDFE generate more effective ADE prediction models compared to automated event-based KDD with incorporated domain knowledge?

In response to RQ1, the step one aKDFE did not result in models with superior prediction performance. Nevertheless, the step two aKDFE (patient-centric transformation) led to models demonstrating higher predictive performance.

RQ2: Does incorporating domain-specific ADE risk scores from a CDSS improve the predictive performance of ADE prediction models?

For RQ2, no statistically significant evidence was found to support the claim that incorporating the CDSS Janusmed risk score improved prediction performance.

## 5.9 Threats to validity and study limitations

Both external and internal threats to validity must be considered. External validity addresses the generalisability of a study's findings to other populations, settings, or contexts. Conversely, internal validity concerns the degree to which the study design and execution accurately reflect the true causal relationship between the variables of interest. This involves examining potential confounding variables and other threats to internal validity, such as selection bias or measurement error [66].

### 5.9.1 Threats to external validity

A key limitation of this study is its external validity. All experiments were conducted on a single project, which restricts the generalisability of the findings to other EHR data registry projects and registry research in general. However, aKDFE utilises low-level features that are not study-case specific, derived from mKDFE, a process demonstrated to generate valuable knowledge [3].

This characteristic suggests a high potential for generalisation. Nevertheless, aKDFE's intrinsic rules might exhibit a high correlation with the specific domain and data used in this study, despite the use of low-level EHR data features. Therefore, future work will investigate this aspect in greater detail.

Furthermore, while a *majority vote* (MV) on a patient level, like the patient-centric data format, could be calculated and evaluated to compare prediction models based on event-based feature sets, this approach has been previously explored in [59], resulting in lower AUROC compared to event-based predictions.

### 5.9.2 Threats to internal validity

As detailed in Section 5.9, internal validity is subject to several potential threats. To mitigate these, this study implemented the following measures: (i) a large sample size was used to minimise the influence of random chance, (ii) rigorous cross-validation was performed across all classifications to ensure robust model evaluation, (iii) training scores were excluded from validation to prevent bias in model selection, and (iv) consistent evaluation metrics were applied across all models for fair comparison.

To prevent selection bias, control patients were sampled by matching on gender, age, index date, and the number of prescribed drugs [58]. These sampling rules were developed by domain experts in ADE research and Janusmed development.

As detailed in Section 2.1, EHR data are primarily intended for clinical use, resulting in registration quality lower than that of *Randomized Controlled Trials* (RCTs). Consequently, quantitative KDD using EHR data necessitates imputation, normalisation, and other transformative preprocessing methods to optimise prediction model generation. These methods inherently introduce potential biases. To mitigate this, we evaluated various preprocessing techniques, cf. Section 3.5.

Furthermore, as outlined in Section 3.6, AUROC was used to evaluate classification performance [55]. However, the choice of performance metric for classification prediction models is a subject of ongoing debate [53, 54, 67-71].

In our experiments, we utilised the native AUROC implementations provided by each classification model. This approach could introduce bias due to variations in AUROC calculation across different models. However, since we consistently applied the same calculation method when comparing AUROC between feature sets, this bias should be minimal, assuming sufficient experiments and robust statistical analysis were performed. These measures were taken to minimise threats to internal validity.

#### 5.9.3 Limitations

While several quantitative classification metrics were collected, as stated in Section 2.10, AUROC was selected as the primary metric for comparing prediction performance between generated models. To statistically validate differences in AUROC, ANOVA F-scores were used. However, it's important to acknowledge that focusing solely on AUROC and ANOVA introduces potential bias, as other metrics and statistical methods might yield different results.

A significant limitation of this study lies in the choices made regarding prediction models, FS methods, and other pipeline components. Specifically, the selection and configuration of recursive models, while potentially valuable in the medical domain, present design, configuration, and interpretation challenges [72].

Furthermore, more advanced and computationally intensive models, such as recursive models [45] and LLMs [26], could be evaluated, as discussed in Section 6.2 and 6.3.

# 6 Related work

The related work section presents relevant research and studies on identifying ADEs in EHR data, with a particular focus on methods like aKDFE.

## 6.1 Automatic feature engineering

While automated feature engineering systems exist, they often rely on intrinsic feature metadata, such as binning HbA1c values into high, normal, and low [32, 73]. However, medical data with temporal information offers significant potential for generating informative features [31]. The aKDFE framework leverages this by supporting temporal operations like "Total number of days between events" and N-grams from ordered event data.

This temporal aspect is often lacking in automated feature engineering systems used in other domains, primarily due to the nature of the domain data [30, 37]. Numerous studies highlight the value of incorporating temporal aspects into feature engineering, demonstrating that while complex to extract, these features are highly informative [31, 74].

## 6.2 Transformation from event to patient-centric data format

In this study, we employed N-grams (*N=1*) to aggregate the engineered and ordered aKDFE features, prioritising a straightforward and explainable automatic method over generating optimally informative features. For prediction models that more effectively leverage temporal information in events or time series, recursive networks are generally preferred [70].

A related study emphasised the importance of patient, clinical, and organisational features when using AI approaches. While recursive models offer significant potential, they are more complex to design and require substantially more training data compared to aKDFE [46].

## 6.3 Entity-centric transformation

While N-grams can represent temporal aspects in event-based data, adaptations of LLMs for time series data, such as Time-LLM or TimeGPT-1, show significant potential for developing more effective prediction models [75, 76]. However, LLMs require considerably more computational resources and training data compared to aKDFE [27].

## 6.4 Predicting ADE in EHR data

Reference [77] reviews studies demonstrating improved ADE prediction models that use unstructured clinical notes and narrative data. The review concludes that NLP and machine learning methods show promise in ADE detection from unstructured EHR data, but standardisation and validation are crucial. Unlike aKDFE, many reviewed methods focused solely on medication events, clinical notes, or other text-based sources. A significant finding of our study is that other EHR data related to patient healthcare history are strongly correlated with ADEs, particularly QT prolongation.

Specifically, language models trained on clinical text, such as ClinicalBERT, have shown superior performance [78].

Furthermore, our study highlights the substantial positive impact of patient-centric transformation on the prediction performance of generated models. Therefore, combining event-based results from NLP models like ClinicalBERT with the patient-centric transformation process in aKDFE (step two) could potentially yield high-performing prediction models.

## 6.5 Use of CDSS with EHR data

In [6], pharmaVISTA and MediQ CDSSs, designed for drug-to-drug-interactions (DDI) detection, initially produced 15.5 alerts per patient. Incorporating medical domain expertise and clinical features like comorbidities and lab results reduced this to 0.8. The study highlighted the necessity of these features for effective CDSSs. Notably, aKDFE automatically identifies and utilises comorbidities, especially diagnoses preceding ADEs (Table 14), which significantly influences prediction model performance.

A review of ADE detection in EHR data [79] revealed that, when employing tabular feature sets, simpler machine learning models, such as gradient boosting trees, outperformed deep learning approaches, which inherently perform automatic feature engineering. This suggests that explicit feature engineering may not be beneficial in this context.

Although aKDFE also employs automatic feature engineering, it produces informative feature sets without the extensive input data requirements of deep learning. This difference can be attributed to aKDFE's pre-model feature engineering, as opposed to the integrated feature engineering of deep learning.

# 7 Conclusion

This section summarises the impact of aKDFE as supported by the presented results. For generating machine learning models to predict ADEs, aKDFE demonstrates greater effectiveness compared to using event-based EHR data alone. However, no statistically significant effect on the AUROC score was observed when Janusmed ADE risk scores and medication route of administration were included in the feature sets.

The patient-centric transformation of event-based feature sets into patient-centric data formatted sets results in a significant enhancement of model prediction performance.

Despite aKDFE's complex iterative feature engineering process, which includes automatic selection of potential compound features, the selected and utilised features can be decoded back to their low-level components with explanatory descriptions.

## 7.1 Future work

Ordered event data shares structural and property similarities with written text and time series data, suggesting the potential applicability of attention mechanisms, as employed in LLMs. Given LLMs' proven effectiveness in NLP, they could be highly effective in identifying patterns within the sequenced features of event feature sets.

Furthermore, developing an automated method for identifying and transforming established domain knowledge into computer-interpretable rules or instructions, and integrating these into the aKDFE framework prior to a new medical registry research project, would be a valuable avenue for exploration.

# Declaration of interest

The authors declare that they have no known competing financial interests or personal relationships that could have appeared to influence the work reported in this paper.

# Acknowledgements

This research received no specific grants from funding agencies in the public, commercial, or not-for-profit sectors.

# Appendix

## Important subfunctions in an machine learning pipeline

Even though all steps in Figure 6 are central and needs to be addressed for valid and sound prediction model generation many published health care related studies lacks key considerations, i.e. class imbalance problems, external validation, and adherence to reporting guidelines [80, 81], see Table A1.

*Table A1 Examples of important subfunctions in an machine learning pipeline.*

| Subfunction | Description | Types, examples |
|---|---|---|
| Imputation | Data from real-world usage, often have a lot of missing values, to adjust several methods are described. The aim is to use as much original information as possible without losing the original implication with methods that fill in the missing data points [82]. | For each feature:<br>○ Mean value<br>○ Max value<br>○ Median value. |
| Normalisation | Transforming data to a common scale. This is crucial because many machine learning algorithms perform better when features are on a similar scale, e.g., between 0 and 1, or standardised to have zero mean and unit variance. | ○ All values in [0-1]<br>○ Zero mean and unified variance. |
| Balancing target class imbalance | Addressing the issue where one class in the dataset has significantly more instances than | ○ Oversampling, e.g., generating synthetic samples for the minority class |
| | another. This can bias the model towards the majority class. | ○ Undersampling, e.g., reducing the number of selected instances of the majority class. |
| Feature selection (FS) | Feature engineering can produce many features, which need to be reduced before applying a predictive model. Otherwise, the number of features has a negative impact due to (i) longer model generation time, and (ii) overfitting of the model leading to a decline in prediction performance on new data [83]. In general, overfitting occurs when too few samples are used in relation to the number of included features [83]. *Different feature selection* (FS) methods can work well depending on the specific feature set. | ○ Recursive feature elimination<br>○ Lasso regulation<br>○ Information gain<br>○ Correlation-based<br>○ No FS, i.e., using all features for classification or prediction [83]. |
| Model selection | Choosing the most appropriate machine learning model for a given problem and dataset. | ○ Decision trees<br>○ Support vector machines<br>○ Neural networks<br>○ Recursive neural network (RNN). |
| Cross-validation (CV) | Divide the dataset into multiple subsets, called folds, of equal size. CV is important in machine learning to avoid model bias [84, 85]. This process is repeated for each fold, ensuring that every part of the dataset is used as a test set exactly once. | ○ K-fold cross-validation: The dataset is divided into 'k' equal-sized folds<br>○ Leave-One-Out cross-validation (LOOCV): A k-fold where k is equal to the number of data points<br>○ Stratified cross-validation: Each fold maintains the same class distribution as the original dataset. |
| Hyperparameter optimisation | Finding the best combination of hyperparameters for a chosen machine learning model to achieve optimal performance. Hyperparameters are settings that control the learning process of the model. | ○ Learning rate<br>○ Number of hidden layers in a neural network<br>○ Tree depth. |

## Implemented machine learning pipeline activities and used methods in the study

In Table A2 detailed information is presented regarding the used methods for each machine learning pipeline activity.

*Table A2 Machine learning pipeline activities and used methods.*

| Machine learning pipeline activity | Method | Parameters | Method description |
|---|---|---|---|
| Data cleaning and transformation | pandas.get_dummies | dummy_na=False | Converts categorical features into binary features, eliminating artificial ordinal relationships. It is best suited for low-cardinality features. |
| Data cleaning and transformation | Mean encoding based on target | None | Reduces the number of generated features when encoded categorical features have high cardinality. Categorical values are replaced by mean. |
| Train validation split | sklearn.model_selection. train_test_split | stratify=target, random_state=42 | Splits the input feature sets by a 0.8 / 0.2 ratio. |
| Imputation | sklearn.preprocessing SimpleImputer | strategy = mean | Imputes missing values with mean value for same feature. |
| Imputation | sklearn.preprocessing SimpleImputer | strategy = median | Imputes missing values with median value for same feature. |
| Normalisation | sklearn.preprocessing MinMaxScaler | None | Numerical feature values are transformed to the range [0,1]. |
| Balancing of target class | imblearn.over_sampling. SMOTE | random_state=42 | Generates synthetic samples of the minority class to balance the dataset. |
| Balancing of target class | imblearn.over_sampling. ADASYN | random_state=42 | Balance imbalanced datasets by generating synthetic samples for the minority class |
| Feature selection | sklearn.feature_selection. SelectKBest | chi2 | Selects the K most relevant features based on how strongly they correlate with the target, as measured by the chi-squared test. |
| Feature selection | sklearn.feature_selection. SelectKBest | f_classif | Selects the "k" features that have the strongest statistical relationship with your target variable based on the ANOVA F-value. |
| Feature selection | sklearn.feature_selection. SelectKBest | mutual_info_classif | Selects the *k* best features based on their mutual information with the target variable. |
| Model selection | sklearn.linear_model. LogisticRegression | max_iter=100, random_state=42, C: [0.001, 0.01, 0.1, 1, 10] | A tool used to build a model that will predict categories, by using a linear function, and a logistic function. |
| Model selection | sklearn.svm.SVF | random_state=42, C: [0.1, 1, 10] | Finds an optimal non-linear hyperplane that best separates data points belonging to different classes. |
| Model selection | sklearn.ensemble. RandomForestClassifier | random_state=42, n_estimators: [100, | Fits multiple decision trees on different |
| | | 200, 300], max_depth: [None, 5, 10] | subtrees for better accuracy and prevention of overfitting. |
| Model selection | sklearn.neighbors. KNeighborsClassifier | n_neighbors: [3, 5, 7] | Classification by vote of the *K* (here *K*=5) nearest neighbours. |
| Model selection | sklearn.svm.LinearSVC | random_state=42, max_iter=1000, C: [0.001, 0.01, 0.1, 1, 10], loss: ['hinge', 'squared_hinge'] | Finds the best straight line (or hyperplane in higher dimensions) to separate data points into different classes. |
| Model selection | sklearn.ensemble. GradientBoostingClassifier | random_state=42, n_estimators: [50, 100, 200], learning_rate: [0.01, 0.1, 1.0], max_depth': [3, 5, 7], subsample: [0.7, 1.0] | Minimise s a loss function by iteratively adding trees that reduce the gradient of the error. |
| Model selection | sklearn.neural_network. MLPClassifier | random_state=42, max_iter=100, hidden_layer_sizes: [(10,), (15,), (20,), (10, 10), (15, 10)], activation: ['relu', 'tanh'], alpha: [0.0001, 0.001, 0.01], learning_rate_init': [0.001, 0.01] | A shallow multilayer classifier that minimise s the log-loss function by evaluating the partial derivates. |
| Model selection | keras.layers.SimpleRNN | Hidden nodes = 16, Regulation = 0.1, Learning rate = 0.1, Momentum = 0.99. | Is used for basic RNN modelling, less effective for capturing long-range dependencies compared to GRU and LSTM. |
| Cross-validation | sklearn.model_selection. GridSearchCV | cv=5, scoring=accuracy, n_jobs=-1 | Performs a thorough search for the best hyperparameters for models using 5-fold cross-validation and accuracy as the evaluation metric. |
| Evaluation of generated models | sklearn.metrics | accuracy_score, precision_score, recall_score, f1_score, roc_auc_score, log_loss, brier_score_loss | Mainly AUROC was evaluated. |

## Feature set in details

Table A3 (1-EVENT), A4 (2-EVENT), and A5 (1-event-aKDFE and 2-event-aKDFE) comprise detailed descriptions of the used feature sets in the study.

To support the evaluation the following feature sets with adhering data structure were used.

*Table A3 Included features for 1-EVENT.*

| Feature name | Feature ID | Feature description |
|---|---|---|
| PATIENT_STUDY AGE DECADE | 1 | Patient age decade at start of study |
| EVENT CONCEPT TYPE ID | 2 | Type of event |
| GENDER | 3 | Patient gender |
| PATIENT AGE AT OBSERVATION | 4 | Patient age at event |
| CENSOR_DATE | 5 | Date of patient death or censoring |
| DRUG USE INDEX GROUP | 6 | Estimated index of patient drug use |
| OBSERVATION_START_DATE | 7 | Date for event |
| VALUE CHAR | 8 | ATC Code or ICD-10 diagnosis code |
| VALUE_DECIMAL | 9 | Event value |
| DRUG DOSAGE FORM CODE | 10 | Drug dosage form type |
| DRUG SUBSTANS ID | 11 | Drug substance ID |

*Table A4 Included features for 2-EVENT.*

| Feature name | Feature ID | Feature description |
|---|---|---|
| PATIENT_STUDY AGE DECADE | 1 | Patient age decade at start of study |
| EVENT CONCEPT TYPE ID | 2 | Type of event |
| GENDER | 3 | Patient gender |
| PATIENT AGE AT OBSERVATION | 4 | Patient age at event |
| CENSOR_DATE | 5 | Date of patient death or censoring |
| DRUG USE INDEX GROUP | 6 | Estimated index of patient drug use |
| OBSERVATION_START_DATE | 7 | Date for event |
| VALUE CHAR | 8 | ATC Code or ICD-10 diagnosis code |
| VALUE_DECIMAL | 9 | Event value |
| DRUG DOSAGE FORM CODE | 10 | Drug dosage form type |
| DRUG SUBSTANS ID | 11 | Drug substance ID |
| ROUTE OF ADMINISTRATION TYPE | 12 | Medication route of administration type |
| DRUG REGISTRATION RISK VALUE | 13 | Risk value for event from Janusmed |

*Table A5 Feature set details for event aKDFE group (1-event-aKDFE and 2-event-aKDFE).*

| Feature name | Feature ID | Feature description |
|---|---|---|
| FEATURE ID | 2 | Feature id for aKDFE generated feature |
| GENDER | 3 | Patient gender |
| PATIENT AGE AT OBSERVATION | 4 | Patient age at feature id event |
| CENSOR_DATE | 5 | Date of patient death or censoring |
| OBSERVATION START DATE | 7 | Date for feature id event |
| VALUE_CHAR | 8 | Character value for feature id |
| VALUE DECIMAL | 9 | Decimal value |

The patient-centric data formatted aKDFE group comprise transformed feature sets from the event aKDFE group. The transformation inferred that the shape of the feature sets went from narrow and long to short and wide.

The patient-centric data formatted feature sets included (i) N-grams (*N=1*) and (ii) sum of feature values from the ordered event aKDFE feature group, see Table A6. The N-grams consisted of two consecutive features, ordered by observation start date.

*Table A6 Details of patient-centric data formatted aKDFE feature sets.*

| Patient-centric data formatted feature set | Type | Naming prefix | N | Feature description |
|---|---|---|---|---|
| 1-PCD-aKDFE | N-gram | NC | 528 | Transformed features (N-grams) |
| 1-PCD-aKDFE | Sum | S | 883 | Transformed features (Sum feature value) |
| 2-PCD-aKDFE | N-gram | NC | 486 | Transformed features (N-grams) |
| 2-PCD-aKDFE | Sum | S | 843 | Transformed features (Sum feature value) |